\documentclass[sigconf,nonacm,authordraft=false]{acmart}
\renewcommand\footnotetextcopyrightpermission[1]{}
\AtBeginDocument{%
  }

\setcopyright{acmlicensed}
\copyrightyear{2018}
\acmYear{2018}
\acmDOI{XXXXXXX.XXXXXXX}

\acmISBN{978-1-4503-XXXX-X/2018/06}

\usepackage[most]{tcolorbox}
\usepackage{stfloats}
\definecolor{promptheader}{RGB}{64, 64, 64}  
\definecolor{promptbody}{RGB}{240, 248, 255} 

\begin{document}
\title{SpotAgent: Grounding Visual Geo-localization in Large Vision-Language Models through Agentic Reasoning}
\thanks{$^*$Both authors contributed equally to this research.}
\thanks{$^\dagger$Work done during an internship at StepFun.}
\thanks{$^\ddagger$Corresponding authors.}
  

\author{
  \makebox[0pt][c]{ 
    \begin{tabular}{c}
    Furong Jia$^{1,*\dagger}$ \quad Ling Dai$^{1,*}$ \quad Wenjin Deng$^{2}$ \quad Fan Zhang$^{1,\ddagger}$ \\ 
      Chen Hu$^{2,\ddagger}$ \quad Daxin Jiang$^{2}$ \quad Yu Liu$^{1}$ \\[0.5ex]
      $^{1}$Peking University \quad $^{2}$StepFun \\
      \texttt{jiafr1802@stu.pku.edu.cn \quad lingdai25@stu.pku.edu.cn}
    \end{tabular}
  }
}






  

\renewcommand{\shortauthors}{Jia et al.}

\begin{abstract}
Large Vision-Language Models (LVLMs) have demonstrated strong reasoning capabilities in geo-localization, yet they often struggle in real-world scenarios where visual cues are sparse, long-tailed, and highly ambiguous. Previous approaches, bound by internal knowledge, often fail to provide verifiable results, yielding confident but ungrounded predictions when faced with confounded evidence. To address these challenges, we propose SpotAgent, a framework that formalizes geo-localization into an agentic reasoning process that leverages expert-level reasoning to synergize visual interpretation with tool-assisted verification. SpotAgent actively explores and verifies visual cues by leveraging external tools (e.g., web search, maps) through a ReAct diagram. We introduce a 3-stage post-training pipeline starting with a Supervised Fine-Tuning (SFT) stage for basic alignment, followed by an Agentic Cold Start phase utilizing high-quality trajectories synthesized via a Multi-Agent framework, aiming to instill tool-calling expertise. Subsequently, the model's reasoning capabilities are refined through Reinforcement Learning. We propose a Spatially-Aware Dynamic Filtering strategy to enhance the efficiency of the RL stage by prioritizing learnable samples based on spatial difficulty. Extensive experiments on standard benchmarks demonstrate that SpotAgent achieves state-of-the-art performance, effectively mitigating hallucinations while delivering precise and verifiable geo-localization.
Our code is available at: \href{https://jiafr1802.github.io/SpotAgent-Paper/}{Project Page}.
\end{abstract}


\begin{teaserfigure}
    \centering
  \includegraphics[width=0.95\textwidth]{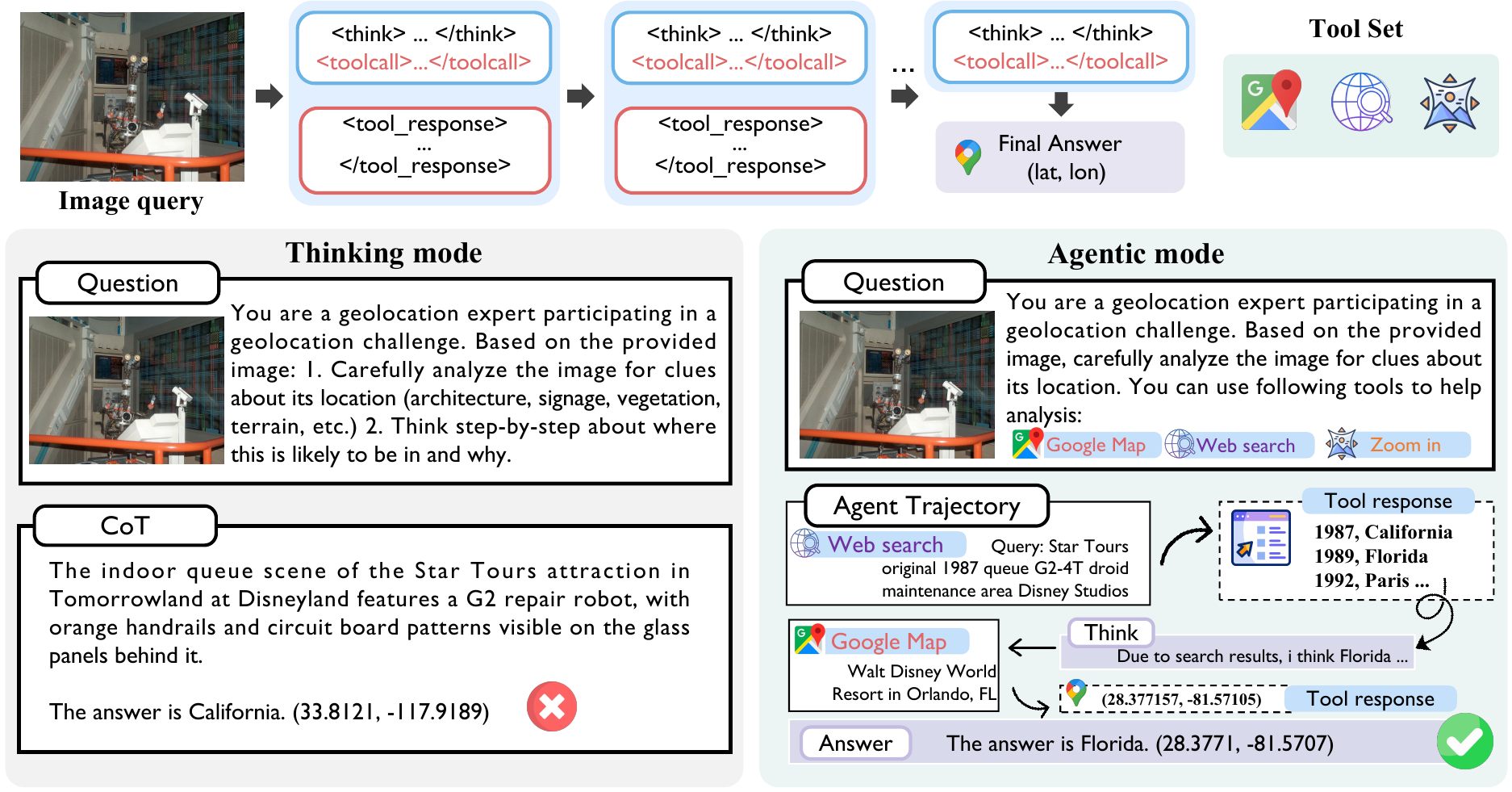}
  \caption{Illustration of SpotAgent operating within a ReAct loop (top) and a comparative of the standard thinking mode versus our agentic mode (bottom).}
  \Description{Enjoying the baseball game from the third-base
  seats. Ichiro Suzuki preparing to bat.}
  \label{fig:teaser}
\end{teaserfigure}


\maketitle

\section{Introduction}
Image geo-localization is the task of predicting the geographic coordinates of a given query image. 
With the rapid growth of visual content on global platforms, this task has become increasingly critical for downstream applications ranging from autonomous navigation~\cite{desouza2002vision} to cultural heritage preservation. 
Traditional approaches typically operate under closed-domain assumptions by either maintaining a candidate database for retrieval~\cite{yang2021cross,zhu2022transgeo,wang2023fine,haas2023learning,xia2025fg} or partitioning the geographical space into fixed grids for classification~\cite{weyand2016planet,seo2018cplanet,muller2018geolocation,pramanick2022world,clark2023we}. 
However, these approaches often compromise the continuous nature of geographical space 
and fail to provide the explicit logical reasoning or interpretability required for complex spatial tasks.

Recent progress in Large Vision-Language Models (LVLMs)~\cite{liu2023visual,bai2025qwen2,chen2024internvl,singh2025openai,comanici2025gemini} has opened new possibilities for reasoning-driven geolocalization. With their massive pre-trained knowledge bases, LVLMs demonstrate impressive capabilities in common scene understanding and foundational geographical reasoning.  
However, real-world geo-localization tasks are inherently dynamic and exhibit a long-tail distribution.
Relying solely on static internal knowledge is insufficient for these challenges \cite{li2025knowledge}, as successful resolution requires the integration of external information to supplement scene understanding.
Consequently, standard LVLMs are susceptible to hallucinations, where they generate plausible but spatially incorrect reasoning paths based on visual stereotypes rather than factual evidence. 
Resulting from a lack of verifiable external knowledge, these models are confined to internal representations, limiting their capacity to handle diverse and challenging tasks.

Inspired by cognitive science findings that active exploration and visual cue extraction are tightly interleaved~\cite{gottlieb2018towards}, we argue that reliable image geo-localization with LVLMs requires moving beyond visual reasoning as the sole source of evidence.
In this paper, we propose SpotAgent, a novel framework that formalizes visual geo-localization into an agent-based decision-making process. 
In contrast to chain-of-thought (CoT) approaches~\cite{wang2025gre} that operate solely within an internal reasoning space, SpotAgent adopts a ReAct-based framework~\cite{yao2022react} that tightly couples reasoning with action, enabling iterative evidence acquisition and refinement.
Equiped with querying tools such as web search and geocoding services~\cite{singh2017evaluating}, the framework orchestrates a ReAct loop that leverages expert-level reasoning to integrate structured visual interpretation with active investigation of external information. This mechanism enables the agent to iteratively filter and structure visual cues from heterogeneous information sources, thereby calibrating predictions against a logically reconstructed factual context.

To transition from static visual thinking to active agentic reasoning, we design a 3-stage post-training pipeline that instills tool-calling skills and expert-level reasoning. 
This is realized through an Agentic Cold Start stage following the initial SFT stage. 
We automate the synthesis of tool-integrated training data by employing a Multi-Agent framework,
where an Observer Agent handles structured visual scene interpretation, and a Tool-Call Agent conducts iterative evidence verification. 
Their collaboration generates evidence-rich, verifiable trajectories that effectively activate the model's agentic potential.
We subsequently incorporate a Reinforcement Learning (RL) stage using Group Relative Policy Optimization (GRPO)~\cite{shao2024deepseekmath} to 
align the agent's reasoning with the objective of achieving fine-grained spatial accuracy.
To enhance learning efficiency, we propose a Spatially-Aware Dynamic Filtering strategy, prioritizing samples within an effective difficulty range by discarding instances that are either trivial or intractable. 

Our key contributions are summarized as follows: 
\begin{itemize}
\item We develop SpotAgent, an image geo-localization agent framework featuring a Multi-Agent data synthesis pipeline that constructs high-quality agentic trajectories with rich evidence, activating tool-use proficiency and systematic external grounding 
without the need for manual labels.
\item We enhance the effectiveness of reinforcement learning for geo-localization agent by introducing Spatially-Aware Dynamic Filtering, a curriculum-based strategy that focus on an optimal difficulty window to guide agent evolution.
\item Extensive experiments demonstrate that SpotAgent achieves state-of-the-art performance on Im2GPS3k among retrieval-free methods, effectively mitigating spatial hallucinations and grounding visual cues in an end-to-end manner.

\end{itemize}

\section{Related Work}

\subsection{Image Geo-localization}
Traditional worldwide geo-localization primarily formulates the task as either a classification problem over discrete grid cells or a cross-modal retrieval task \cite{weyand2016planet, vo2017revisiting}. Classification-based approaches discretize the Earth's surface into predefined regions and train models to assign each image to a specific cell \cite{seo2018cplanet, clark2023geodecoder}. Early efforts such as PlaNet \cite{weyand2016planet} pioneered this paradigm, which was later enhanced by combinatorial partitioning in CPlaNet \cite{seo2018cplanet} and hierarchical models in ISNs \cite{muller2018geolocation}. Another dominant paradigm is retrieval-based geo-localization, which matches query images against massive GPS-tagged galleries \cite{vo2017revisiting, zhou2024img2loc}.
Some research have introduced transformer-based architectures like Translocator \cite{pramanick2022world}, contrastive frameworks such as GeoCLIP \cite{vivanco2023geoclip} and PIGEON \cite{haas2024pigeon}.
Beyond classification methods and retrieval-based methods, diffusion-based generative models have also been proposed to address the geo-localization problem \cite{dufour2025around}.

\subsection{Reasoning-driven Image Geo-localization with LVLMs}

The emergence of LVLMs has enabled a shift from simple visual recognition to reasoning-driven geo-localization \cite{li2025globe, wu2026geor}. 
Frameworks including G3 \cite{jia2024g3} and GeoRanker \cite{jia2025georanker} introduce an early form of retrieval-augmented grounding (RAG) for LVLM-based geo-localization, where external image--GPS galleries or textual knowledge are incorporated to support reasoning in a RAG-style manner. 
By leveraging the extensive geographical knowledge encoded in foundation models such as Qwen2.5-VL~\cite{bai2025qwen2}, researchers have adopted CoT paradigms to justify coordinate predictions through natural language deduction \cite{ye2024where}. 
GRE Suite \cite{wang2025gre} further demonstrates that modeling geo-localization as a structured reasoning task can significantly improve prediction accuracy. 
In particular, GLOBE \cite{li2025globe} introduces policy optimization by utilizing verifiable entities as reward signals, and Geo-R \cite{wu2026geor} employs RL with format-oriented rewards to ensure strictly structured outputs.
Existing LVLM approaches treated geo-localization task as a closed-world reasoning task, where the model must infer locations solely from internal representations without the ability to interact with external knowledge sources.

\subsection{Agentic Reasoning for Visual Task}

The evolution of LLM agents has shifted the paradigm from Chain-of-Thought reasoning to the synergy of reasoning and acting. Foundations like ReAct~\cite{yao2022react} and Toolformer~\cite{schick2023toolformer} demonstrated that interleaving reasoning traces with task-specific actions enables models to break the limitations of static black box, allowing them to dynamiclly interface with external environments to gather information. Building on this, early multimodal frameworks such as MM-REACT~\cite{yang2023mm} and ViperGPT~\cite{suris2023vipergpt} extended this capability to the visual domain by orchestrating specialized vision models or synthesizing code to resolve visual queries. External visual tools serve as a perceptual bridge that unleashes the VLM's latent grounding capabilities, enabling rigorous logical reasoning on fine-grained visual details~\cite{yang2023set}.
Recently, domain-specific agents have emerged to address vertical tasks. 
Refocus~\cite{fu2025refocus} employs visual tools for chart understanding.
PixelCraft~\cite{zhang2025pixelcraft} achieves high-fidelity visual reasoning on structured images by integrating fine-tuned pixel-level grounding with specialized visual tools.
GeoVista~\cite{wang2025geovista} enables web-augmented geo-localization but lacks the capacity for coordinate-level quantification, often failing to yield precise geographic results.

\section{Method}

\subsection{Problem Formulation}
\label{subsec:problem_formulation}
\begin{figure}[h]
  \centering
  \includegraphics[width=\linewidth]{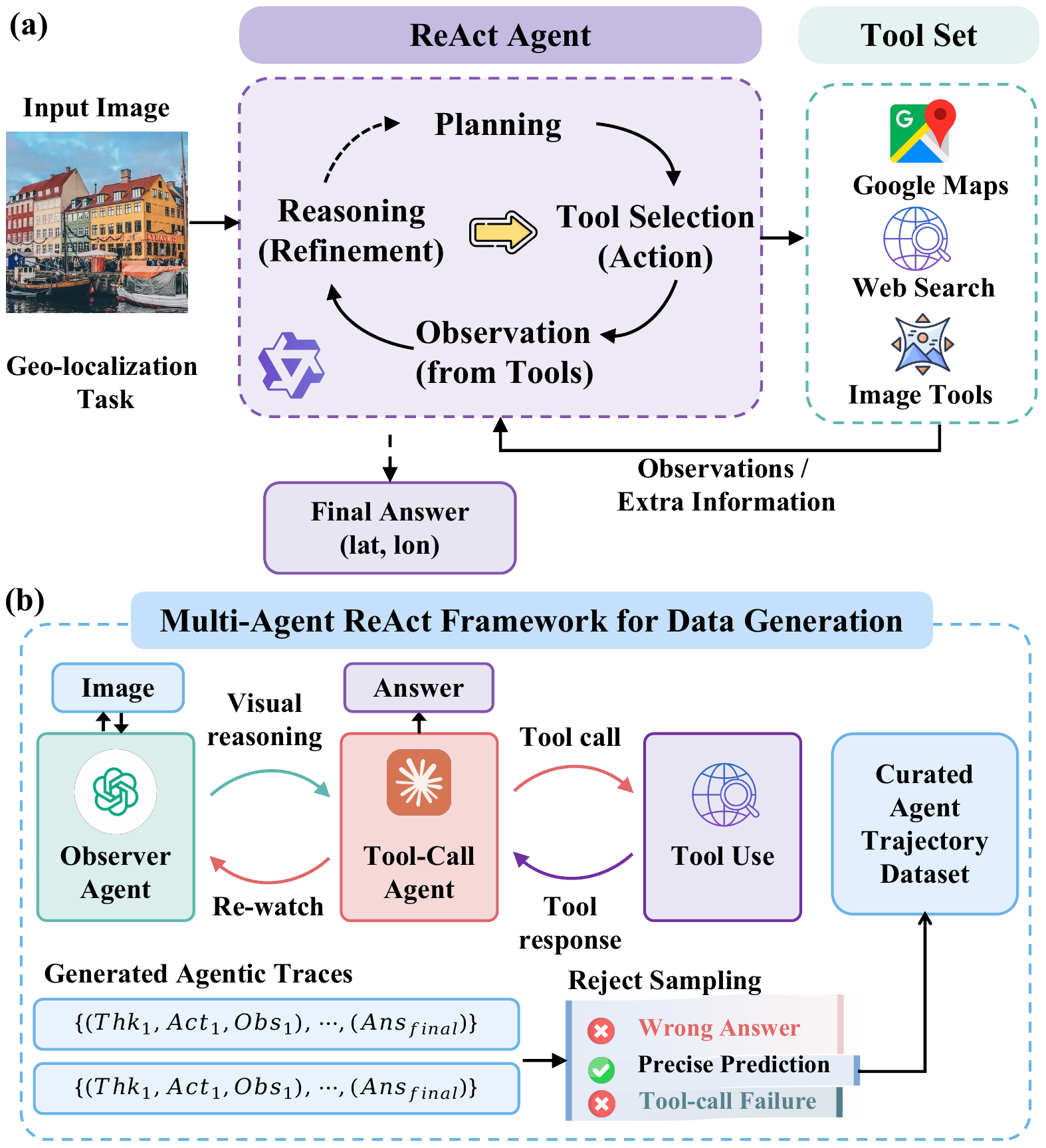}
  \caption{Overview of the proposed framework. (a) Inference Pipeline: SpotAgent operates as a ReAct loop, formulating geo-localization as a sequential decision-making process to interact with external tools. (b) Data Generation Pipeline: A Multi-Agent framework is employed for synthesizing high-quality training trajectories.}
  \Description{TODO}
  \label{fig:agent_overview} 
\end{figure}

Traditional visual geo-localization methods typically formulate the task as $y = f_\theta(x)$,
where $x$ is the input image and $y$ denotes the geographic coordinates. 
LVLM-based methods employing CoT reasoning for geo-localization fundamentally rely on their internal weights $\theta$ to derive answers. 
We formalize visual geo-localization not as a static reasoning task, but as a sequential decision-making process (Figure~\ref{fig:agent_overview}a) that interleaves visual reasoning traces with active tool execution.
Specifically, we formulate the agent-based visual geo-localization task as a Partially Observable Markov Decision Process (POMDP), defined by the tuple $\mathcal{M} = (\mathcal{S}, \mathcal{A}, \mathcal{O}, \mathcal{T}, \mathcal{R})$.

\textbf{State Space ($\mathcal{S}$):} 
At time step $t$, the state $s_t \in \mathcal{S}$ represents the comprehensive context available to the agent. 
Formally, $s_t$ is defined as the tuple $\{I_q, H_t\}$, where $I_q$ denotes the initial visual input (i.e., the query image) and $H_t = [m_1, m_2, \dots, m_{t-1}]$ represents the history of interaction messages. 
This history sequence $H_t$ accumulates the agent's past reasoning steps, tool invocation commands, and the corresponding observations returned by the environment.

\textbf{Action Space ($\mathcal{A}$):} The agent selects actions from a predefined toolset $\mathcal{A} = \mathcal{A}_{tools} \cup \mathcal{A}_{text}$.
\begin{itemize}
    \item $\mathcal{A}_{tools}$: A set of exploratory actions grounded in external tools, including:
    \begin{itemize}
        \item \texttt{GeoCoding($p$)}: Converts a textual place name $p$ into candidate coordinates, implemented by Google Maps.
        \item \texttt{WebSearch($q$)}: Queries a search engine to retrieve information for external knowledge.
        \item \texttt{ImageTool($img$)}: Executes the zoom-in operation.
    \end{itemize}
    \item $\mathcal{A}_{\text{text}}$: The text generation action where the agent outputs a natural language sequence representing its internal reasoning process. This allows the agent to analyze the visual content, plan the next steps, or summarize information from history before executing a tool or making a final decision.
\end{itemize}

\textbf{Observation Space ($\mathcal{O}$):} At step $t=0$, the observation is the query image $I_q$. At subsequent steps $t > 0$, the observation $o_t$ corresponds to the return value of the executed tool $a_{t-1}$, including search snippets, map coordinates, or cropped images.

\textbf{Policy ($\pi$):} The agent functions as a policy $\pi_\theta(a_t | h_t)$, where $h_t = \{I_q, a_1, o_1, \dots, a_{t-1}, o_{t-1}\}$ represents the thinking and interaction history. The goal is to learn a policy that maximizes the probability of the correct location in a restricted number of interaction steps.

Formally, unlike standard CoT paradigm, which models the probability of the answer $y$ conditioned solely on the image $I_q$ and internal reasoning $r$:
\begin{equation}
    \pi_{CoT}(y | I_q) =  P(y | r, I_q) P(r | I_q)
\end{equation}
Our agentic formulation conditions the prediction on the dynamic evidence gathered through the trajectory $h_t$:
\begin{equation}
\begin{split}
    \pi_{Agent}(y | I_q, \mathcal{A}_{tools}) = P(y | h_T), \quad \\ 
    \text{where} \quad h_T = \{I_q, (a_0, o_0), \dots, (a_T, o_T)\}
\end{split}
\end{equation}

To ensure efficiency and prevent infinite loops, we impose a strict budget on the agent's interaction steps:
\begin{equation}
    \text{subject to} \quad t \leq T_{\text{max}}, \quad \forall a_t \in h_T \cup \mathcal{A}_{tools}
\end{equation}
where $T_{\text{max}}$ denotes the maximum allowable budget for total tool invocations (set to $6$ in our implementation) throughout the agentic reasoning process.

\subsection{Agentic Approach to Image Geolocalization}
\label{sec:agent}

\subsubsection{\textbf{Multi-Agent ReAct Framework for Data Generation}}
\label{subsec:agent-data}
To synthesize a high-quality post-training dataset featuring expert-level reasoning and active tool use, we design a Multi-Agent ReAct framework that decouples visual perception from strategic planning. 
As illustrated in Figure~\ref{fig:agent_overview}b, this architecture consists of two agents: an \textit{Observer Agent} based on a VLM, and a \textit{Tool-Call Agent} based on an LLM.
This approach augments single VLMs in zero-shot settings, leveraging LLMs to enhance the rigorous, multi-step inferencing capabilities required for complex verification~\cite{qiao2024prism,you2023idealgpt,chen2023large}.

\paragraph{\textbf{Observer Agent}.}
The \textit{Observer Agent} performs structured, hierarchical reasoning to interpret the query image $I$ or specific zoomed-in regions. Rather than providing a flat description, it generates a multi-level visual interpretation $d_{vis}$ that systematically decomposes the scene. 
This coarse-to-fine taxonomical analysis spans from macro-scale environmental features, such as architectural styles and terrain, to micro-scale semantic clues, including specific text and infrastructure. This structured visual interpretation ensures that subsequent information inquiry is grounded in a rigorous visual foundation:
\begin{equation}
d_{vis} = \pi_{obs}(I, P, b_{zoom})
\end{equation}
where $P$ denotes the task-specific prompt guiding a structured interpretation, and $b_{zoom} \in \mathcal{B} \cup \{\emptyset\}$ represents the optional bounding box generated by the visual zoom-in tool (where $b_{zoom} = \emptyset$ implies default global observation).

\paragraph{\textbf{Tool-Call Agent}.}
The \textit{Tool-Call Agent} relies on $d_{vis}$ and the interaction history $h_t$ to interact with the tools and environment. At each step $t$, it synthesizes a reasoning thought ($Thk_t$) to analyze the current situation, followed by a decision action ($Act_t$):
\begin{equation}
    (Thk_t, Act_t) \sim \pi_{plan}(\cdot | h_t, d_{vis})
\end{equation}
The generated action $Act_t$ falls into two categories: it is either a tool invocation ($Act_t \in \mathcal{A}_{tools}$) to gather more evidence, or a termination action where the agent concludes the reasoning to output the final answer $Ans_{final}$.
Specifically, $\mathcal{A}_{tools}$ comprises both external knowledge discovery utilities (e.g., \texttt{WebSearch}) and active visual modules (e.g., \texttt{ImageTool}). Triggering the latter prompts the Observer Agent to re-watch the image context, such as zooming in on a signboard, to support the subsequent reasoning step.

\paragraph{\textbf{Rejection Sampling for Trajectory Curation.}}
To ensure the quality of the training data, we employ a rigorous rejection sampling strategy. We denote a generated trajectory as:
\begin{equation}
\tau = \{(Thk_1, Act_1, Obs_1), \dots, (Ans_{final})\}
\end{equation}
A trajectory $\tau$ is accepted into the curated dataset $\mathcal{D}_{curated}$ only if it satisfies two criteria:
\begin{itemize}
    \item \text{Tool-Call validity:} Every action $a_{tool} \in \tau$ must be executable, including valid tool-call format and successful response, ensuring the agent learns correct tool usage.
    \item \text{Prediction precision:} The final predicted coordinates $\hat{y}$ must fall within a strict distance threshold $\delta = 5km$ from the ground truth $y^*$:
    \begin{equation}
        \text{Dist}(\hat{y}, y^*) < \delta
    \end{equation}
\end{itemize}
\noindent In practice, the rejection sampling filters $\sim$20k raw trajectories down to $\sim$6,000 validated samples; we refer to this collection $\mathcal{D}_{curated}$ as the \textbf{SpotAgenticCoT-6k} dataset.

\paragraph{\textbf{Agentic Trajectory Data Formatting.}}
To facilitate structured learning and clear delineation between reasoning, actions, and observations, each trajectory in \textbf{SpotAgenticCoT-6k} is formatted using specialized tags. The reasoning process is encapsulated within \texttt{<think>\allowbreak...\allowbreak</think>} blocks. Tool interactions follow a strict request-response pattern:
\begin{itemize}
    \item \text{Tool invocations:} Actions are wrapped in \texttt{<tool\_call>\allowbreak...\allowbreak</tool\_call>} tags. Inside these tags, the agent specifies the tool name and its required parameters in JSON format.
    \item \text{Tool response:} The execution results from the tools are wrapped in \texttt{<tool\_response>\allowbreak...\allowbreak</tool\_response>} tags.
\end{itemize}

\subsubsection{\textbf{Post-Training Framework of SpotAgent}}
\begin{figure*}[t]  
  \centering
  \includegraphics[width=0.95\linewidth]{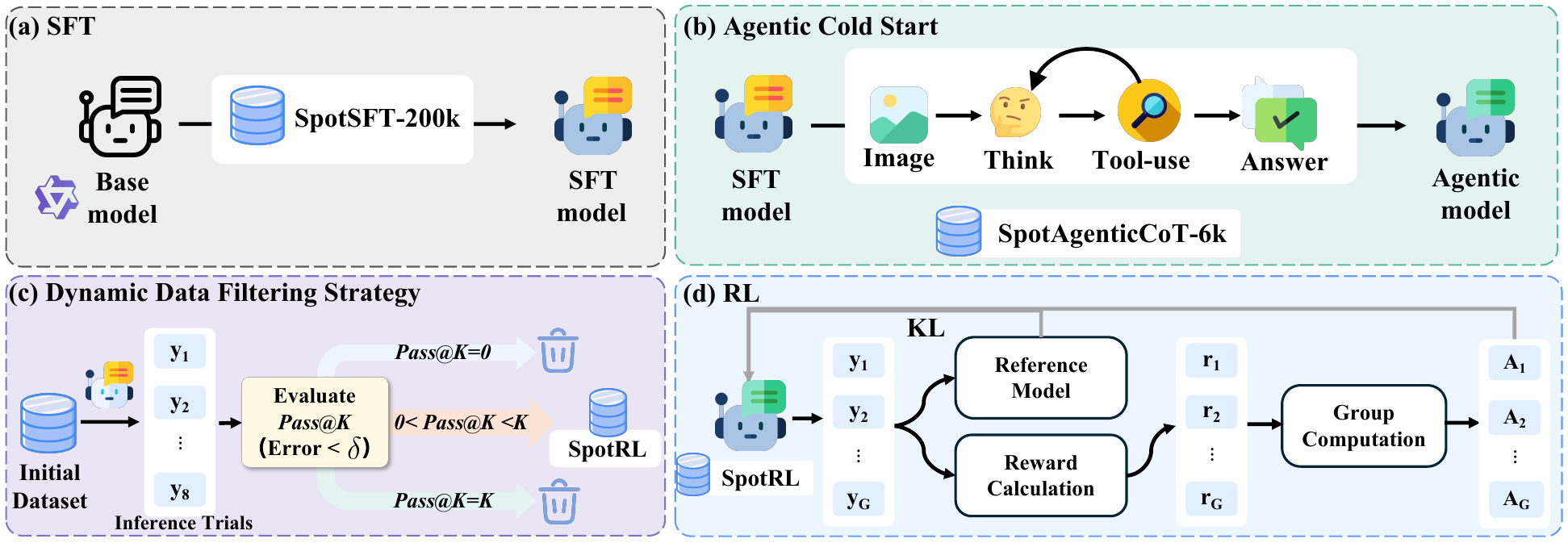} 
  \caption{Post-training framework of SpotAgent. (a) The base model is first aligned to the geo-localization task through supervised fine-tuning. (b) Agentic capabilities are then activated via an agentic cold start stage, where the model learns to perform agentic reasoning with tool interactions. (c) A dynamic data filtering strategy selects learnable samples to improve RL optimization efficiency. (d) The model's reasoning capabilities is further refined through RL using the filtered dataset.}
  \Description{TODO}
  \label{fig:figure2}
\end{figure*}
We design a 3-stage post-training pipeline for SpotAgent as illustrated in Figure~\ref{fig:figure2}. This framework proceeds in three stages: establishing basic capability via SFT, activating tool-use skills through an agentic cold start, and finally refining reasoning precision via RL.

\paragraph{\textbf{Supervised Fine-Tuning (SFT)}}
To establish foundational capabilities for visual geo-localization, we first perform SFT on the base LVLM (Figure~\ref{fig:figure2}a). We randomly sample 5\% of the MP16-Pro dataset ($\sim$200k samples) and organize the image-coordinate pairs into a Visual QA format, constructing a dataset we term \textbf{SpotSFT-200K}. 
The automated construction of this dataset provides an efficient, low-cost pathway to inject basic world knowledge and align the model with general geo-localization instruction formats.

\paragraph{\textbf{Agentic Cold Start}}
We subsequently implement an agentic cold start stage to instill tool-calling capabilities (Figure~\ref{fig:figure2}b), utilizing the high-quality trajectories in \textbf{SpotAgenticCoT-6k} dataset synthesized from Section~\ref{subsec:agent-data}.
During this stage, we supervise only the assistant's responses; specifically, the training loss is computed exclusively on the reasoning traces and tool invocations while masking the user inputs and tool observations.

\subsection{Reinforcement Learning with Spatially-Aware Dynamic Filtering}
\label{sec:rl}

Subsequent to supervised fine-tuning and the agentic cold start phase, we utilize reinforcement learning to specifically refine the agent's reasoning capabilities in a reasoning mode without tool invocation (Figure~\ref{fig:figure2}d).

\subsubsection{\textbf{Dynamic Data Filtering Strategy for RL}}
To maximize data and learning efficiency, we construct our RL training dataset, \textbf{SpotRL} on the model’s edge of competence~\cite{zhang2025interplay}, through a dynamic filtering process applied on the initial SFT dataset. 
We prioritize training on instances that provide the most effective learning signals by focusing on samples that are neither fully mastered nor entirely intractable (Figure~\ref{fig:figure2}c).

We perform $K=8$ inference trials for each image using the SFT model. Let $\text{Pass}@K$ denote the number of trials where the prediction error falls within a specific distance threshold $\delta$. 
We categorize the filtered data into distinct data regimes defined by the distance thresholds $\delta \in \{1\text{km}, 25\text{km}, 200\text{km}, 750\text{km}\}$, retaining only learnable samples satisfying $0 < \text{Pass}@K < K$:
\begin{itemize}
    \item Removing the Intractable Regime ($\text{Pass}@K = 0$): Discarding samples where the model fails consistently.
    \item Removing the Mastered Regime ($\text{Pass}@K = K$): Discarding samples that are already solved in every trial to avoid diminishing returns.
\end{itemize}

\subsubsection{\textbf{Spatially-Aware Curriculum Learning}}
We implement a two-stage curriculum learning strategy utilizing different data regimes of \textbf{SpotRL}:
\begin{itemize}
    \item Phase I. We initialize training with the strict data regimes ($\delta \in \{1\text{km}, 25\text{km}\}$), focusing on samples where the model already achieves relatively high precision but requires further reasoning refinement.
    \item Phase II. Upon convergence of Phase I, we expand the training set to include looser data regimes filtered by looser thresholds ($\delta \in \{200\text{km}, 750\text{km}\}$), incorporating harder samples where the model previously achieved only coarse localization.
\end{itemize}

\subsubsection{\textbf{Geodesic Distance-based Reward Design}}
\label{subsubsec:reward}
We design a linear hierarchical reward function for RL stage.
The reward $r$ for a predicted coordinate $\hat{y}$ given ground truth $y$ and geodesic error $d(\hat{y}, y)$ is defined as a piecewise linear decay:
\begin{equation}
  R(d) =
  \begin{cases}
    1.0 & \text{if } d < 1 \text{ km} \\
    1.0 - \frac{0.25}{24}(d - 1) & \text{if } 1 \le d < 25 \text{ km} \\
    0.75 - \frac{0.55}{175}(d - 25) & \text{if } 25 \le d < 200 \text{ km} \\
    0 & \text{otherwise}
  \end{cases}
\end{equation}
where $d(\hat{y}, y)$ denotes the great-circle distance between the prediction and the ground truth.

Unlike prior works~\cite{wu2026geor,wang2025gre} that rely on explicit rewards to enforce output formatting, we omit such auxiliary signals. 
This is because the geodesic distance reward serves as an inherent structural constraint: the computation of $R(d)$ strictly depends on the successful parsing of valid coordinates. Consequently, the model is implicitly incentivized to adhere to the specified format to maximize its expected return.

\subsubsection{\textbf{Optimization via Group Relative Policy Optimization (GRPO)}}

For each image input $x$ in our curated batch, we sample a group of $G$ outputs $\{y_1, y_2, \dots, y_G\}$ from the current policy $\pi_\theta$, where $G=8$. We then compute the reward $r_i$ for each output $y_i$ using the reward function defined in Section~\ref{subsubsec:reward}. 
GRPO computes the advantage $A_i$ for the $i$-th output by normalizing its reward against the peer outputs in the same group. This serves as a dynamic, instance-specific baseline:
\begin{equation}
A_i = \frac{r_i - \mu(\{r_1, \dots, r_G\})}{\sigma(\{r_1, \dots, r_G\}) + \epsilon}
\end{equation}
where $\mu$ and $\sigma$ are the mean and standard deviation of the rewards within the group, and $\epsilon$ is a small constant for numerical stability. 
Then the policy is optimized by maximizing the surrogate objective, constraining the update with a KL-divergence penalty to prevent deviation from the reference SFT model $\pi_{ref}$:
\begin{equation}
\begin{split}
    \mathcal{L}_{GRPO}(\theta) = \mathbb{E}_{x \sim D, y \sim \pi_\theta} \bigg[ &\frac{1}{G} \sum_{i=1}^G \min \left( \rho_i A_i, \operatorname{clip}(\rho_i, 1-\epsilon, 1+\epsilon) A_i \right) \\
    &- \beta_{KL} D_{KL}(\pi_\theta || \pi_{ref}) \bigg]
\end{split}
\end{equation}
where $\rho_i = \frac{\pi_\theta(y_i|x)}{\pi_{old}(y_i|x)}$ is the importance sampling probability ratio. 

\section{Experiments}
\subsection{Main Results}
\subsubsection{\textbf{Performance on Image Geo-localization}}
We assess our model's geo-localization accuracy in comparison with established methods from the literature. Specifically, we conduct our evaluations on two standard benchmarks: Im2GPS3k~\cite{hays2008im2gps} and YFCC4K. 
The comparative results are reported in Table~\ref{tab:geoloc_results}.
Accuracy is measured by calculating the geodesic distance between predicted and ground-truth coordinates, reporting the percentage of samples that fall within a series of distance thresholds (1 km, 25 km, 200 km, 750 km, and 2500 km).
Given the known sensitivity of our base VLM (Qwen2.5-VL-7B-Instruct) to prompting strategies, we present both our reproduction results (`Our Eval.') and the figures reported in prior publication~\cite{li2025globe} (`Prior Work') for the Qwen2.5-VL-7B baseline in Table~\ref{tab:geoloc_results} to ensure a comprehensive comparison.
Notably, our method outperforms the competitive retrieval-free approaches on Im2GPS3k, achieving 14.12\% and 40.36\% accuracy at the 1km and 25km thresholds, respectively.

\begin{table*}[t]
\centering
\caption{Comparison of geo-localization performance on Im2GPS3k and YFCC4K benchmarks. We report the percentage (\%) of test images localized within various distance thresholds. The best results in retrieval-free methods are highlighted in \textbf{bold}.}
\label{tab:geoloc_results}

\setlength{\tabcolsep}{3.5pt}

\begin{tabular}{@{}lc ccccc | ccccc@{}}
\toprule
 & & \multicolumn{5}{c}{\textbf{Im2GPS3k}} & \multicolumn{5}{c}{\textbf{YFCC4K}} \\
\cmidrule(lr){3-7} \cmidrule(l){8-12}
\textbf{Methods} & \textbf{Venue} & \textbf{1km} & \textbf{25km} & \textbf{200km} & \textbf{750km} & \textbf{2500km} & \textbf{1km} & \textbf{25km} & \textbf{200km} & \textbf{750km} & \textbf{2500km} \\
\midrule

\multicolumn{12}{l}{\textit{Retrieval-based Methods}} \\
{[L]}kNN, $\sigma =4$ ~\cite{vo2017revisiting} & ICCV'17 & 7.2 & 19.4 & 26.9 & 38.9 & 55.9 & 2.3 & 5.7 & 11.0 & 23.5 & 42.0 \\
Img2Loc ~\cite{zhou2024img2loc}  & SIGIR'24 & 15.34 & 39.83 & 53.59 & 69.7 & 82.78 & 19.78 & 30.71 & 41.4 & 58.11 & 74.07 \\
PIGEON ~\cite{haas2024pigeon} & CVPR'24 & 11.3 & 36.7 & 53.8 & 72.4 & 85.3 & 10.4 & 23.7 & 40.6 & 62.2 & 77.7 \\
G3 ~\cite{jia2024g3} & NeurIPS'24 & 16.65 & 40.94 & 55.56 & 71.24 & 84.68 & - & - & - & - & - \\
\midrule
\multicolumn{12}{l}{\textit{Generalist Vision-Language Models (VLMs)}} \\
Qwen2.5-VL-7B ~\cite{bai2025qwen2} & Our Eval. & 3.97 & 25.48 & 43.65 & 64.29 & 77.90 &  1.90 & 13.03 & 24.21 & 40.30 & 57.28 \\
Qwen2.5-VL-7B ~\cite{bai2025qwen2} & Prior Work~\cite{li2025globe} & 8.58 & 32.53 & 43.11 & 58.93 & 72.37 & - & - & - & - & - \\
\midrule
\multicolumn{12}{l}{\textit{Retrieval-free Methods}} \\
PlaNet ~\cite{weyand2016planet} & ECCV'16 & 8.5 & 24.8 & 34.3 & 48.4 & 64.6 & 5.6 & 14.3 & 22.2 & 36.4 & 55.8 \\
CPlaNet ~\cite{seo2018cplanet} & ECCV'18 & 10.2 & 26.5 & 34.6 & 48.6 & 64.6 & 7.9 & 14.8 & 21.9 & 36.4 & 55.5 \\
ISNs ~\cite{muller2018geolocation} & ECCV'18 & 3.2 & 9.6 & 14.3 & 25.1 & 43.9 & 6.5 & 16.2 & 23.8 & 37.4 & 55.0 \\
Translocator ~\cite{pramanick2022world} & ECCV'22 & 7.6 & 20.3 & 27.1 & 40.7 & 63.3 & 8.4 & 18.6 & 27.0 & 41.1 & 60.4 \\
GeoDecoder ~\cite{clark2023we} & ICCV'23 & 5.7 & 10.3 & 21.4 & 28.9 & 38.6 & \textbf{10.3} & \textbf{24.4} & 33.9 & 50.0 & 68.7 \\
GeoCLIP ~\cite{vivanco2023geoclip} & NeurIPS'23 & 14.11 & 34.47 & 50.65 & 69.67 & 83.82 & 9.59 & 19.31 & 32.63 & 55.00 & \textbf{74.69} \\
GRE suite ~\cite{wang2025gre} & NeurIPS'25 & 11.3 & 35.3 & 51.7 & 69.3 & 85.7 & - & - & - & - & - \\
GLOBE ~\cite{li2025globe} & NeurIPS'25 & 9.84 & 40.18 & 56.19 & 71.45 & 82.38 & - & - & - & - & - \\

\midrule
\textbf{SpotAgent (Ours)} & - & \textbf{14.12} & \textbf{40.36} & \textbf{57.80} & \textbf{73.43} & \textbf{85.75} & 7.30 & 21.52 & \textbf{36.18} & \textbf{55.00} & 70.77\\
\bottomrule
\end{tabular}
\end{table*}

\subsubsection{\textbf{Agentic Behavior Boost Performance on Long-Tail Data}}

\begin{figure}[h]
  \centering
  \includegraphics[width=\linewidth]{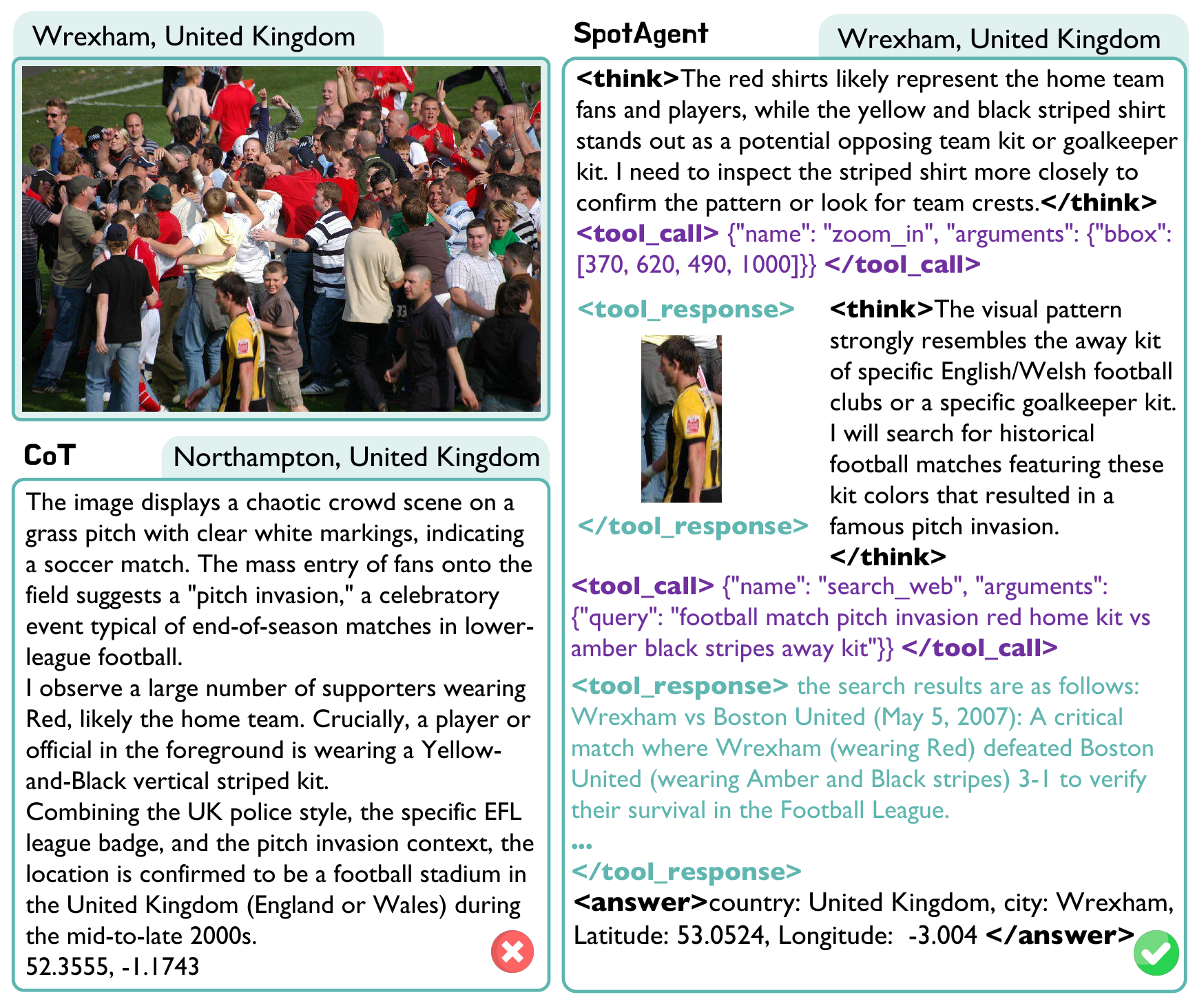}
  \caption{Reasoning comparison: CoT Mode vs. Agentic Mode on an Im2GPS3k benchmark image.}
  \Description{TODO}
  \label{fig:case_Im2GPS3k} 
\end{figure}

Long-tail scenarios in geo-localization are characterized by the absence of salient landmarks, requiring the connection of obscure visual details like ephemeral events or specific text with vast external knowledge. To demonstrate superior performance in such cases, we qualitatively compare SpotAgent against a variant restricted from tool access, which effectively functions as a standard CoT model. As illustrated in Figure \ref{fig:case_Im2GPS3k}, while the CoT mode correctly identifies the general context of a pitch invasion, it fails to recognize specific team identifiers and consequently mislocates the event.
In contrast, SpotAgent executes a structured reasoning strategy: it first employs visual zoom-in to isolate the specific ``amber-and-black'' striped kit, then utilizes web search to cross-reference this detail with historical data, pinpointing the specific event as the 2007 match between Wrexham and Boston United. Following this, the agent determines the venue as the home team's stadium to obtain the final coordinates. This workflow demonstrates how agentic information discovery effectively resolves long-tail ambiguity, converting obscure visual details into decisive localization evidence.

\begin{table}[htbp]
  \small 
  \caption{Results of The Street View Text Dataset}
  \label{tab:street_view}
  \begin{tabular}{l ccccc} 
    \hline
    \rule{0pt}{12pt} 
    & \multicolumn{5}{c}{\textbf{Accuracy (\%)}} \\
    \cline{2-6} 
    \rule{0pt}{12pt}
    \textbf{Methods} & \textbf{1km} & \textbf{25km} & \textbf{200km} & \textbf{750km} & \textbf{2500km} \\
    \hline
    \rule{0pt}{12pt}
    No Tool       & 9.22  & 28.82 & 38.90 & 52.45 & 76.95 \\
    Tool-assisted & \textbf{19.65} & \textbf{38.60} & \textbf{50.18} & \textbf{60.00} & \textbf{82.81} \\
    \hline
    \rule{0pt}{12pt}

  \end{tabular}
\end{table}

To further investigate the model's performance on ``semantic long-tail'' scenarios, where geo-localization relies on decoding specific textual entities (e.g., shop signs) rather than global visual landmarks, we extend our evaluation to the Street View Text Dataset~\cite{wang2010word}. The quantitative results presented in Table~\ref{tab:street_view} demonstrate that SpotAgent significantly outperforms the variant without tools.

\begin{figure}[h]
  \centering
  \includegraphics[width=\linewidth]{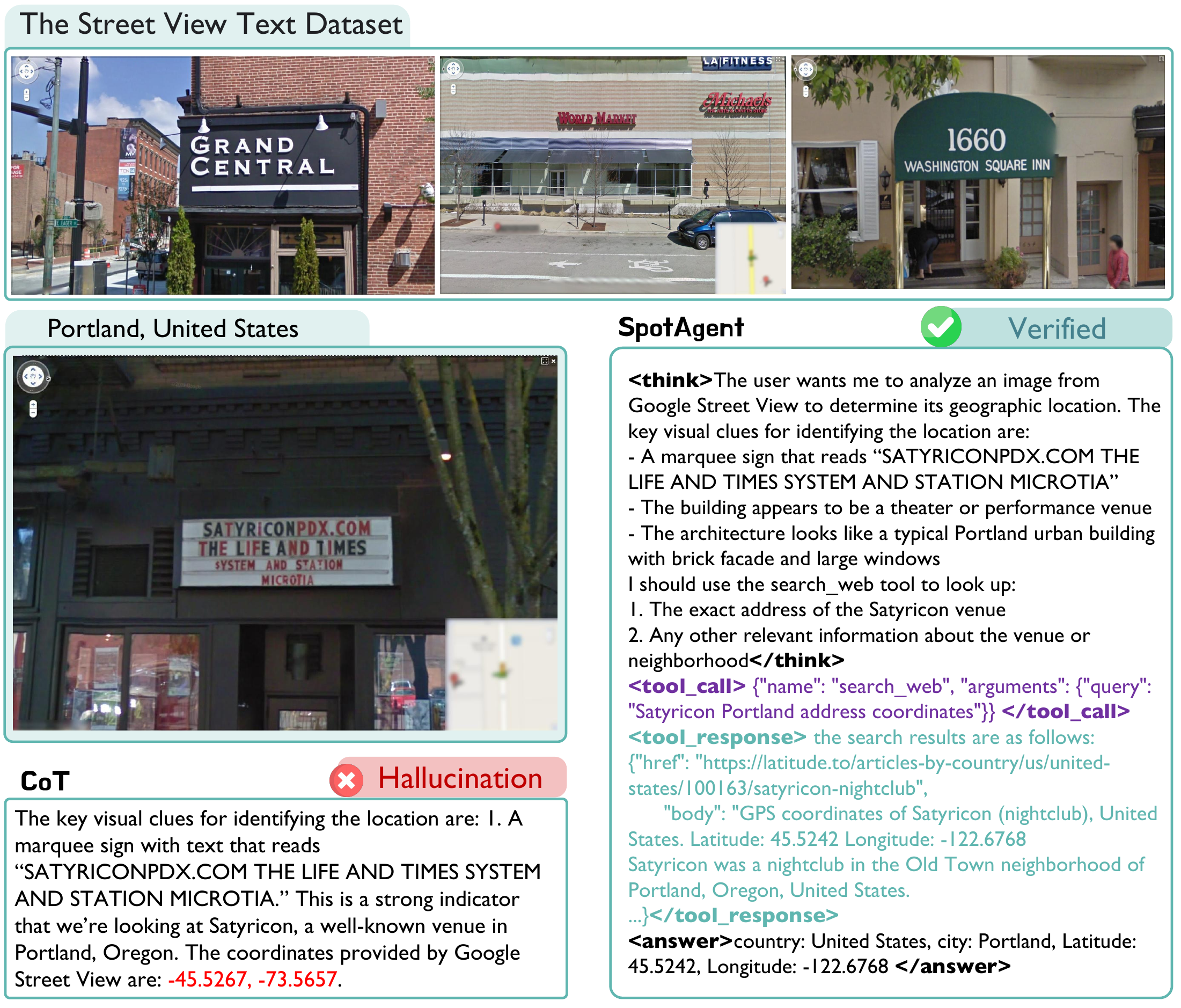}
  \caption{Reasoning comparison: CoT Mode vs. Agentic Mode on an image from Street View Text Dataset.}
  \Description{TODO}
  \label{fig:case_streetview} 
\end{figure}

Qualitative analysis reveals that while standard CoT processes can successfully identify key visual clues such as business names, they often fall into spatial hallucinations by generating plausible but incorrect coordinates due to a lack of verification (Figure~\ref{fig:case_streetview}a). In contrast, the agentic mode enables the model to proactively invoke web search tools to cross-reference extracted textual information with external knowledge(Figure~\ref{fig:case_streetview}b). By transforming visual cues into verifiable search queries, the agent achieves far higher credibility and precision, proving that an active, tool-assisted framework is essential for reliable geo-localization in complex, information-dense environments.

\subsubsection{\textbf{Agent Trajectory Analysis}} 

\begin{figure}[h]
  \centering
  \includegraphics[width=0.95\linewidth]{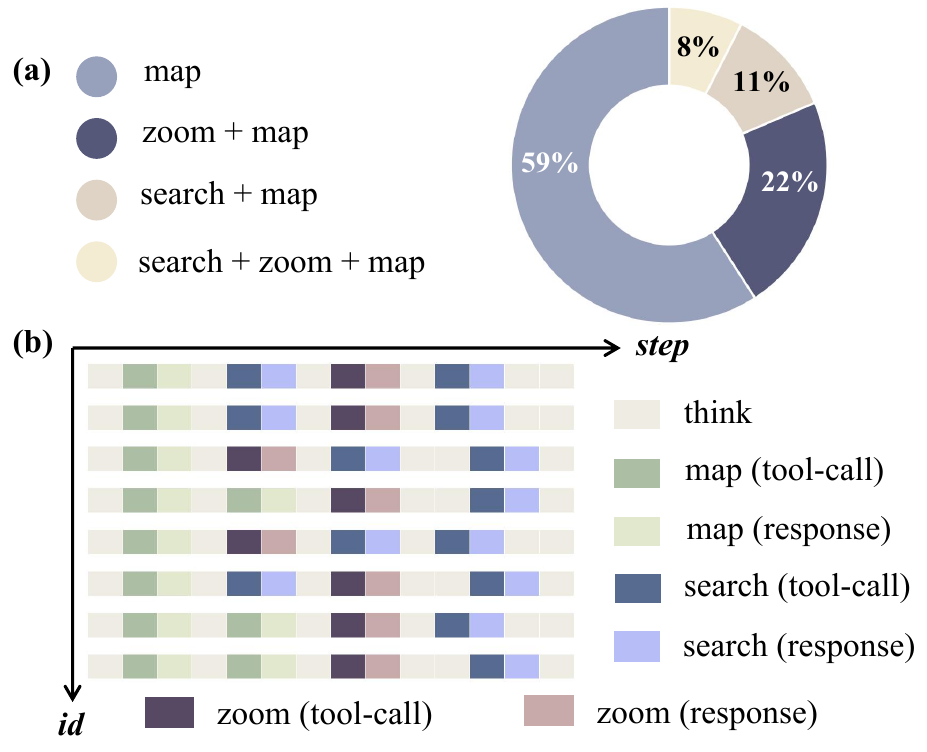}
  \caption{Analysis of Agent Behaviors on Im2GPS3k. (a) Distribution of tool usage combinations. (b) Visualization of the step-wise reasoning trajectories.} \label{fig:trajectory_analysis}
\end{figure}

To gain deeper insights into the behavioral dynamics of SpotAgent, we conduct a granular analysis of the inference trajectories on the Im2GPS3k benchmark, as visualized in Figure~\ref{fig:trajectory_analysis}.

\noindent \textbf{Analysis of Tool Usage.} Figure~\ref{fig:trajectory_analysis}a illustrates the distribution of tool utilization patterns. 
A dominant portion (59\%) of trajectories relies solely on the \texttt{GeoCoding} tool, suggesting that the agent is capable of recognizing the landmark or place name directly from the image, requiring the tool to refine this semantic prediction into exact coordinates.
41\% of cases involve combinations of \texttt{ImageTool} and \texttt{WebSearch}. This indicates that the agent triggers active visual re-examination (\texttt{ImageTool}) or external exploration (\texttt{WebSearch}) only when the visual evidence and internal knowledge is insufficient, rather than blindly invoking all tools.

\noindent \textbf{Step-wise Trajectory Visualization.} Figure~\ref{fig:trajectory_analysis}b provides a microscopic view of the decision-making process across representative trajectories. We observe a clear, structured interleaving of the \textit{Think-Act-Observe} cycle. 
These samples with consistent path lengths are randomly chosen to visualize the distinct usage patterns of the three available tools.

\subsubsection{\textbf{Necessity of Agentic Cold Start for Effective Tool Interaction}}
\label{sec:acs_analysis}

\begin{figure}[h]
    \centering
    \includegraphics[width=1.0\linewidth]{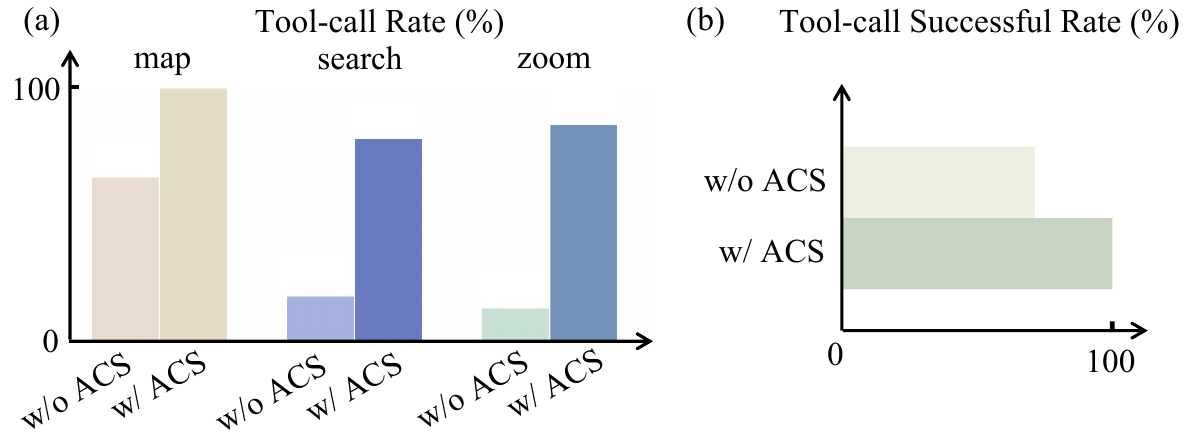}
    \caption{Impact of Agentic Cold Start (ACS) on tool interactions on the Im2GPS3k benchmark. (a) \textbf{Tool-call Rate}: The frequency of calling specific tools. Note that the y-axis is plotted on a \textbf{log scale}. (b) \textbf{Tool-call Success Rate}: The proportion of generated tool calls that are valid and executable.}
    \label{fig:tool_analysis}
\end{figure}

To validate the necessity of the Agentic Cold Start (ACS) strategy, we conducted a quantitative analysis of the model's tool-use behavior on the Im2GPS3k benchmark. We compared the performance of the model trained with and without the ACS stage. The results are illustrated in Figure~\ref{fig:tool_analysis}.

\paragraph{\textbf{Activation of Tool-Use Awareness.}}
As shown in Figure~\ref{fig:tool_analysis}a, we first examined the Tool-call Rate, which measures how frequently the model attempts to invoke external tools during reasoning. 
Without ACS, the model exhibits an extremely low tendency to utilize tools. 
In contrast, the integration of ACS activates the model's agentic awareness, resulting in a substantial increase in tool usage across all categories.

\paragraph{\textbf{Enhancement of Execution Reliability.}}
Beyond the frequency of usage, the quality of the tool calls is critical. Figure~\ref{fig:tool_analysis}b illustrates the Tool-call Success Rate, defined as the percentage of tool calls that are valid, parseable, and executable. 
The results demonstrate that without ACS, even when the model attempts to use a tool, it frequently suffers from formatting errors. 
This stage effectively bridges this gap by grounding the model in correct tool-use trajectories, ensuring high execution reliability. 


\subsection{Ablation Studies}

\subsubsection{\textbf{Efficacy of Tool-Augmented Reasoning}}

\begin{table}[htbp]
  \small
  \centering
  \caption{Impact of different tool configurations. Abbreviations: WS (Web Search), VZ (Visual Zoom), GC (Geo Coding).}
  \label{tab:ablation_tools}
  \begin{tabular}{l ccccc}
    \hline
    \rule{0pt}{12pt}
    & \multicolumn{5}{c}{\textbf{Accuracy (\%)}} \\
    \cline{2-6}
    \rule{0pt}{12pt}
    \textbf{Tool Config} & \textbf{1km} & \textbf{25km} & \textbf{200km} & \textbf{750km} & \textbf{2500km} \\
    \hline
    \rule{0pt}{12pt}
    Web Search Only           & 12.58 & 38.10 & 54.75 & 70.80 & 82.35 \\
    Visual Zoom Only          & 11.53 & 36.93 & 54.22 & 71.13 & 83.74 \\
    Geo Coding Only        & 12.10 & 37.03 & 52.31 & 68.77 & 80.62 \\
    WS \& VZ          & 12.97 & 38.82 & 56.30 & 71.88 & 82.88 \\
    GC \& VZ          & 13.68 & 38.07 & 52.49 & 67.00 & 78.11 \\
    WS \& GC           & 13.65 & 38.47 & 54.09 & 69.10 & 80.25 \\
    Full Toolset       & \textbf{14.12} & \textbf{40.36} & \textbf{57.80} & \textbf{73.43} & \textbf{85.75} \\
    \hline
  \end{tabular}
\end{table}

To investigate the impact of external tool usage and different tool configurations on the agent's geo-localization capabilities, we conducted an ablation analysis by varying the available tool configurations. 
Table ~\ref{tab:ablation_tools} reports the performance of different tool configurations on the Im2GPS3k benchmark across standard distance thresholds.
Utilizing single-modality tools such as ``Visual Zoom Only'' or ``Geo Coding Only'' provides a functional baseline, and the integration of multiple tools begins to unlock stronger reasoning capabilities.
The full toolset configuration, which enables the agent to dynamically leverage visual details, textual search capabilities, and precise geocoding, yields the highest performance across all reported metrics.
These results demonstrate that the agent's ability to synergize diverse sources of external knowledge is fundamental to achieving precise geo-localization.

\subsubsection{\textbf{Effectiveness of Spatially-Aware Dynamic Filtering}}

\begin{table}[htbp]
  \small
  \centering
  \caption{Comparison of different data sampling strategies during the RL stage.}
  \label{tab:ablation_filtering}
  \begin{tabular}{c ccccc}
    \hline
    \rule{0pt}{12pt}
    & \multicolumn{5}{c}{\textbf{Accuracy (\%)}} \\
    \cline{2-6}
    \rule{0pt}{12pt}
    \textbf{RL Strategy} & \textbf{1km} & \textbf{25km} & \textbf{200km} & \textbf{750km} & \textbf{2500km} \\
    \hline
    \rule{0pt}{12pt}
    Random Sampling   & 13.78 & 39.54 & 56.76 & 72.71 & 84.75 \\
    Dynamic Filtering  & \textbf{14.12} & \textbf{40.36} & \textbf{57.80} & \textbf{73.43} & \textbf{85.75} \\
    \hline
  \end{tabular}
\end{table}

To evaluate the effectiveness of our proposed Spatially-Aware Dynamic Filtering strategy during the RL stage, we compared it against a baseline random sampling approach on the Im2GPS3k benchmark. 
As shown in Table ~\ref{tab:ablation_filtering}, the random sampling method selects an equivalent number of training samples from the same dataset pool without considering sample difficulty. 
Our dynamic filtering strategy consistently outperforms random sampling. 

\subsubsection{\textbf{Step-wise Analysis of Post-Training Strategies}}

\begin{table}[htbp]
  \small
  \centering
  \caption{Ablation study on progressive post-training stages.}
  \label{tab:ablation_stages}
  \begin{tabular}{l ccccc}
    \hline
    \rule{0pt}{12pt}
    & \multicolumn{5}{c}{\textbf{Accuracy (\%)}} \\
    \cline{2-6}
    \rule{0pt}{12pt}
    \textbf{Training Stages} & \textbf{1km} & \textbf{25km} & \textbf{200km} & \textbf{750km} & \textbf{2500km} \\
    \hline
    \rule{0pt}{12pt}
    Qwen2.5-VL-7B        & 3.97 & 25.48 & 43.65 & 64.29 & 77.90 \\
    + SFT \& Cold Start      & 10.03 & 33.67 & 50.62 & 66.88 & 80.77 \\
    + RL                & \textbf{14.12} & \textbf{40.36} & \textbf{57.80} & \textbf{73.43} & \textbf{85.75} \\
    \hline
  \end{tabular}
\end{table}

We ablate the progressive impact of each post-training stage, and the results on the Im2GPS3k benchmark are shown in Table~\ref{tab:ablation_stages}.
For a fair comparison of internal knowledge acquisition, the base Qwen2.5-VL-7B and the SFT \& Cold Start models were evaluated in a standard, tool-free setting.
The initial SFT \& Cold Start phase significantly boosts the model's fundamental localization capabilities, raising the 1km accuracy from 3.97\% to 10.03\%.
Subsequent RL optimizes reasoning logic to yield substantial gains.


\section{Conclusions}

In this paper, we presented SpotAgent, a novel framework that formalizes visual geo-localization from a static reasoning task into an active decision-making process. By equipping Large Vision-Language Models with external tools and a ReAct-style reasoning loop, our approach effectively overcomes parametric knowledge boundaries and spatial hallucinations by grounding visual evidence in verifiable facts. 
To systematically instill these capabilities, we introduced a progressive post-training pipeline comprising Supervised Fine-Tuning, an Agentic Cold Start phase utilizing synthesized Multi-Agent trajectories, and a Reinforcement Learning stage. Crucially, we proposed a Spatially-Aware Dynamic Filtering strategy to optimize learning efficiency by prioritizing samples based on their difficulty. Extensive experiments on standard benchmarks demonstrate that SpotAgent achieves state-of-the-art performance, validating the efficacy of active tool use and iterative verification in achieving precise localization in complex real-world scenarios.
Discussions of the limitations of our framework are provided in Appendix \ref{limit}.

%
\bibliographystyle{ACM-Reference-Format}
\bibliography{sample-base}

%
\appendix
\raggedbottom
\section{Agent Implementation Details}

\subsection{Prompts}
To ensure consistency across different training stages and evaluation settings, we adopt a unified prompt template for all LVLM-based geo-localization experiments. This prompt is designed to guide the model toward systematic reasoning about geographic visual cues while explicitly enforcing a structured Thought–Action pattern for tool usage. 

\begin{tcolorbox}[
    enhanced,
    title={Prompt Template},
    fonttitle=\normalfont,
    colbacktitle=promptheader,
    fontupper=\small,           
    coltitle=white,
    colback=promptbody,
    colframe=black,
    boxrule=0.8pt,
    arc=2mm,
    left=6pt, right=6pt, top=6pt, bottom=6pt,
]

You are a helpful assistant. Your task is to determine the geographic location of an image through systematic visual analysis and external verification.

\vspace{6pt}
\textbf{Reasoning Requirement}

Before calling any tool, you must first reason step-by-step and explicitly state:

\begin{itemize}
\item What visual evidence you have observed
\item What information is still missing
\item Why a specific tool is needed
\end{itemize}

This reasoning must be written inside \texttt{<think> ... </think>} tags.

\vspace{6pt}
\textbf{Available Tools}

The agent can invoke the following tools:

\begin{itemize}
\item \textbf{Image Zoom In Tool}: Crop and zoom into discriminative regions.
    \begin{itemize}
        \item \textit{Parameters}: \texttt{bbox\_2d} [x1, y1, x2, y2] — Defines the area for fine-grained inspection.
    \end{itemize}
\item \textbf{Web Search Tool}: Search the web to verify hypotheses derived from visual evidence.
    \begin{itemize}
        \item \textit{Parameters}: \texttt{query} — The specific search string used to verify visual hypotheses or retrieve landmark coordinates.
    \end{itemize}
\item \textbf{Google Maps Geocoding Tool}: Convert textual locations into geographic coordinates.
    \begin{itemize}
        \item \textit{Parameters}: \texttt{address} — The target place name or street address identified during the visual reasoning stage.
    \end{itemize}
\end{itemize}

Tool calls must be placed inside \texttt{<tool\_call> ... </tool\_call>} tags.

\vspace{6pt}
\textbf{Final Answer Format}

When ready to answer, the agent must:

\begin{itemize}
\item Provide the final answer inside \texttt{<answer> ... </answer>}
\end{itemize}

The final answer must include:

\begin{itemize}
\item Country
\item City
\item Latitude (decimal format)
\item Longitude (decimal format)
\end{itemize}

Latitude and longitude must always be valid numeric values and cannot be null.

\end{tcolorbox}


\subsection{Agent Tool Infra}
\label{appendix:tool_infra}

To ensure modularity and extensibility, we implemented the tool-use capability of SpotAgent following the Model Context Protocol (MCP) design pattern. This allows the agent to interact with external services through a unified, standardized interface.

\begin{itemize}
    \item \textbf{Geocoding Tool Implementation:} The \texttt{maps\_geocode} tool is encapsulated as a dedicated MCP server wrapping the Google Maps Geocoding API. By standardizing the input schema (address strings) and output schema (formatted coordinates with confidence scores). 
    We choose Google Maps for geocoding to guarantee comprehensive global coverage across diverse geographic regions.
    
    \item \textbf{Web Search Tool Implementation:} We implement the search tool with a pluggable backend architecture. The unified search interface normalizes search results, extracting titles, snippets, and URLs, into a fixed JSON structure. 
    During our development, we tested multiple search tool providers, including Tavily and YDC, to verify the agent's robustness across different information acquisition sources.
    
    \item \textbf{Visual Tool Implementation:} Implemented via a local Python-based image processing logic.
\end{itemize}

To verify the robustness of SpotAgent across different web search tools and information acquisition sources, we extended our evaluation to YDC, complementing the Tavily-based results reported in our main experiments.
Notably, YDC offers a more generous free tier, making it a cost-effective alternative for researchers reproducing our work. 
We replaced the Tavily backend with YDC in the MCP implementation while keeping all other parameters fixed. The comparative results on Im2GPS3k are shown in Table~\ref{tab:search_backend_comparison}.

\begin{table}[h]
\centering
\caption{\textbf{Robustness across search tools.} Comparison between Tavily and YDC. The agent demonstrates backend-agnostic robustness. Baseline is the result without using search tool.}
\label{tab:search_backend_comparison}
\resizebox{0.95\linewidth}{!}{
\begin{tabular}{lccccc}
\toprule
 & \multicolumn{5}{c}{\textbf{Accuracy (\%)}} \\ 
 \cmidrule(l){2-6} 
\textbf{Search Backend} & \textbf{1km} & \textbf{25km} & \textbf{200km} & \textbf{750km} & \textbf{2500km} \\
\midrule
Tavily & 14.12 & \textbf{40.36} & \textbf{57.80} & \textbf{73.43} & \textbf{85.75} \\
YDC & \textbf{14.41} & 39.56 & 56.53 & 72.27 & 83.53 \\
\midrule
Baseline & 13.68 & 38.07 & 52.49 & 67.00 & 78.11 \\
\bottomrule
\end{tabular}
}
\end{table}

The performance profile remains mainly consistent across backends. 
The YDC-backed agent achieves 14.62\% accuracy at the strict 1km threshold, slightly outperforming the Tavily baseline. 
Although minor fluctuations occur at coarser levels (e.g., 200km), the model remains superior to the no-search baseline, demonstrating the robustness of our search implementation.

\subsection{Training Data Example}
The example below illustrates a typical training sample from our SpotAgenticCoT-6k dataset. 
Each training instance contains a primary scene image and, if triggered by the agent, a subsequent zoomed-in crop generated via the image zoom-in tool. Accompanying these visual inputs is a complete, multi-step trajectory consisting of interleaved reasoning traces, tool invocations, and 
final geo-coordinates.
The agentic trajectories shown below are synthesized by our Multi-Agent ReAct framework described in Section~\ref{subsec:agent-data}.
Detailed search engine responses are omitted from the example below due to length constraints; however, they are processed by the agent in their entirety. Supplementary annotations are incorporated to elucidate the logical transitions across multiple interaction steps.

\begin{tcolorbox}[
    enhanced,
    title={Example Agentic Trajectory from SpotAgenticCoT-6k},
    fonttitle=\normalfont,
    colbacktitle=promptheader,
    fontupper=\small,           
    coltitle=white,
    colback=promptbody,
    colframe=black,
    boxrule=0.8pt,
    arc=2mm,
    left=6pt, right=6pt, top=6pt, bottom=6pt,
]

\noindent
\begin{tabular}{@{}l p{0.8\textwidth}@{}}
    \textbf{Input:} & \texttt{5dabcda6-94f2-474f-9cc2-c52466b2ae20\_0.jpg} \\
                    & \texttt{5dabcda6-94f2-474f-9cc2-c52466b2ae20\_1.jpg}
\end{tabular}

\vspace{3pt}
\textbf{Step 1 — Initial Visual Reasoning}


\texttt{<think>}
\begin{itemize}
    \item \textbf{Macro-scale Environment}: Urban East Asian streetscape with a dense metropolitan layout.
    \item \textbf{Municipal Infrastructure}:  Modern glass advertising panels resembling standardized bus-stop. Sidewalks with tactile paving patterns typical of Chinese municipal engineering.
    \item \textbf{Semantic \& Mobile Clues}: A yellow city bus, consistent with Beijing's public transit fleet from the 2000s.
    \item \textbf{Inference}: The confluence of specialized infrastructure and transport markers strongly suggests China.
\end{itemize}
\texttt{</think>}

\vspace{3pt}
\textbf{Step 2 — Zoom-in Tool Invocation}

\texttt{<think>} I need to zoom into a region that may contain discriminative text. \texttt{</think>}

\texttt{<tool\_call>} image\_zoom\_in\_tool(bbox\_2d=[350,560,422,636]) \texttt{</tool\_call>}

\textit{Tool response reveals text fragment: “P1.CN”.}

\vspace{3pt}
\textbf{Step 3 — Web Search Tool Invocation}

\texttt{<think>}  
The “.CN” suffix confirms China. I will search for contextual evidence related to Beijing buses and street design.  
\texttt{</think>}

\texttt{<tool\_call>} search\_web(``Beijing yellow bus 2000s city bus fleet'') \texttt{</tool\_call>}

Search results confirm:

\begin{itemize}
\item Beijing’s characteristic yellow buses
\item Modern bus shelter design consistent with Beijing
\end{itemize}

\vspace{3pt}
\textbf{Step 4 — Coordinate Searching}

\texttt{<think>}  
I will now search for coordinates of major commercial districts in Beijing.  
\texttt{</think>}

\texttt{<tool\_call>} search\_web(``Wangfujing Beijing coordinates'') \texttt{</tool\_call>}

\vspace{5pt}
\textbf{Final Grounded Answer}

\texttt{<answer>}  

Country: China \\
City: Beijing \\
Latitude: 39.9067 \\
Longitude: 116.4032  

\texttt{</answer>}

\end{tcolorbox}

\subsection{Details of Multi-Agent Data Generation}

We leverage the Multi-Agent ReAct Framework for automated data synthesis (Section~\ref{subsec:agent-data}). In our implementation, we orchestrate a pipeline of state-of-the-art LLMs to generate samples:

\begin{itemize}
    \item \textbf{Observer Agent}: We employ GPT-5 as the core observer, leveraging its advanced multimodal perception to perform deep semantic analysis of complex visual scenes.
    
    \item \textbf{Tool-call Agent}: For the interactive reasoning and tool-invocation stages, we employ Claude 4 Opus. This model is selected for its superior capability in structured reasoning and accurate tool calling.
\end{itemize}

\section{Training Details}

\subsection{Training Setup}

All experiments are conducted on a single node equipped with 8$\times$ NVIDIA H800 (80GB) GPUs interconnected via NVLink. 

\noindent\textbf{RL framework and system integration.}
For the reinforcement learning stage, we adopt the VERL framework, a post-training system designed for large language models with verifiable reward signals. VERL provides modular interfaces for actor–reference policy management, rollout orchestration, and reward evaluation. We integrate VERL with PyTorch FSDP for memory-efficient distributed training and vLLM as the inference engine for online rollouts.

\noindent\textbf{Distributed training and precision.}
Training is performed with mixed precision (bf16) and FSDP-style full sharding of parameters and optimizer states. Gradient checkpointing is enabled to support multimodal context processing. Gradient accumulation is used to match the effective global batch sizes across 8 GPUs.

\noindent\textbf{Inference engine for rollouts and evaluation.}
All online rollouts during RL are executed using vLLM with tensor model parallelism. The rollout engine is shared between training-time sampling and validation-time evaluation to minimize distribution shifts. For each prompt, the policy samples $n=8$ candidate trajectories during GRPO optimization.


\subsection{Hyperparameters}

\noindent\textbf{Notation.}
train\_batch\_size denotes the effective number of sequences per optimizer step after gradient accumulation across 8 GPUs. ppo\_mini\_batch\_size specifies the per-update minibatch size for optimization. max\_prompt\_length and max\_response\_length are both set to 4096 tokens.

\noindent\textbf{GRPO configuration.}
We use GRPO as the policy optimization algorithm with group size $n=8$ rollouts per prompt. KL regularization is enabled with coefficient $\lambda_{kl}=0.001$ using the low-variance KL formulation. Entropy regularization is disabled. The actor learning rate is set to $1\times10^{-6}$.

\noindent\textbf{Batching and parallelism.}
The training batch size is 256, with a PPO minibatch size of 128. Each GPU processes micro-batches of size 8 during forward passes. vLLM rollout uses tensor model parallel size 2 with GPU memory utilization capped at 0.8.




\noindent\textbf{Optimization details.}
We train for a single epoch over the dataset with validation every 10 steps and checkpointing every 100 steps. All experiments use identical decoding configurations during rollout and evaluation to ensure fair comparison.
\begin{table}[h]
\centering
\small
\caption{Training hyperparameters of SpotAgent (RL stage).}
\begin{tabular}{c c}
\toprule
\textbf{Hyper-parameter} & \textbf{Value} \\
\midrule
Learning Rate & $1\times10^{-6}$ \\
Optimizer & AdamW \\
KL Coefficient $\lambda_{kl}$ & 0.001 \\
KL Type & low\_var\_kl \\
Rollout Group Size $n$ & 8 \\
Train Batch Size & 256 \\
PPO Mini-batch Size & 128 \\
Micro-batch per GPU & 8 \\
Tensor Parallel Size (vLLM) & 2 \\
GPU Memory Utilization (vLLM) & 0.8 \\
Max Prompt Length & 4096 \\
Max Response Length & 4096 \\
Precision & bf16 \\
Gradient Checkpointing & Enabled \\
FSDP Sharding & Full parameter + optimizer state \\
Validation Interval & 10 steps \\
Checkpoint Interval & 100 steps \\
 
\bottomrule
\end{tabular}
\end{table}

\subsection{Details of Dynamic Data Filtering Strategy}

\begin{figure}[h]  
    \centering
    \includegraphics[width=\linewidth]{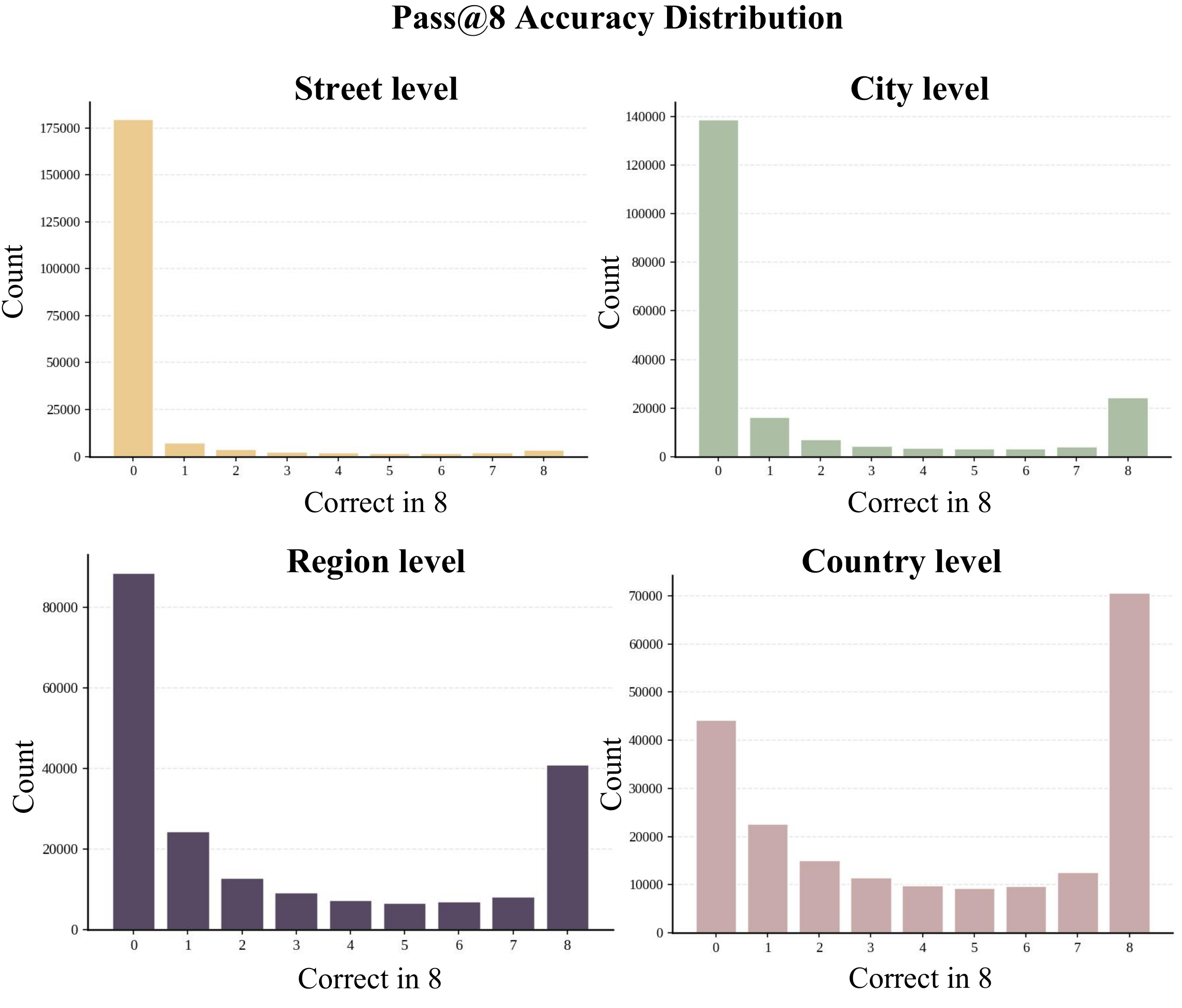}
    \caption{Pass@k across different spatial thresholds.}
    \label{fig:pass}
\end{figure}

Following the Dynamic Data Filtering Strategy (Section~\ref{sec:rl}), we apply $\text{Pass}@k\space ( k=8)$ to filter training data. Figure~\ref{fig:pass} illustrates the accuracy during filtering across four granularity levels defined by the threshold $\delta$: street ($1\text{km}$), city ($25\text{km}$), region ($200\text{km}$), and country ($750\text{km}$).

\begin{table}[htbp]
\centering
\small 
\caption{Statistics of the dynamic data filtering process across four granularity levels. We exclude samples that are either too trivial ($Pass@k=8$) or intractable ($Pass@k=0$ ) to focus training on the most informative instances.}
\label{tab:data_filtering}
\setlength{\tabcolsep}{5pt} 
\begin{tabular}{@{}l ccc@{}}
\toprule
\textbf{Threshold} & \textbf{$Pass@k=8$} & \textbf{$Pass@k=0$} & \textbf{Filtered} \\ \midrule
Street (1km)       & 3,675 (1.78\%)         & 179,665 (87.17\%)          & \textbf{22,766 (11.05\%)} \\
City (25km)        & 24,394 (11.84\%)       & 138,948 (67.42\%)          & \textbf{42,764 (20.75\%)} \\
Region (200km)     & 41,027 (19.91\%)       & 88,712 (43.04\%)           & \textbf{76,367 (37.05\%)} \\
Country (750km)    & 70,729 (34.32\%)       & 44,317 (21.50\%)           & \textbf{91,060 (44.18\%)} \\ \bottomrule
\end{tabular}
\end{table}

As summarized in Table \ref{tab:data_filtering}, our Spatially-Aware Dynamic Filtering strategy identifies varying proportions of informative samples across different spatial scales. At the Street level (1km), a vast majority (87.17\%) of the initial synthesized trajectories are deemed intractable ($Pass@k=0$), reflecting the extreme difficulty of fine-grained localization. Conversely, at broader scales like the Country level (750km), more samples (34.32\%) become trivial ($Pass@k=8$). 
By filtering out these two extremes, we retain a high-quality subset, ranging from 11.0\% to 44.18\% of the total pool that lies within the optimal difficulty range for different stage of reinforcement learning. 

\section{Detailed of Datasets and Benchmarks}
\label{sec:appendix_datasets}

\begin{figure*}[b]  
    \centering
    \includegraphics[width=0.85\linewidth]{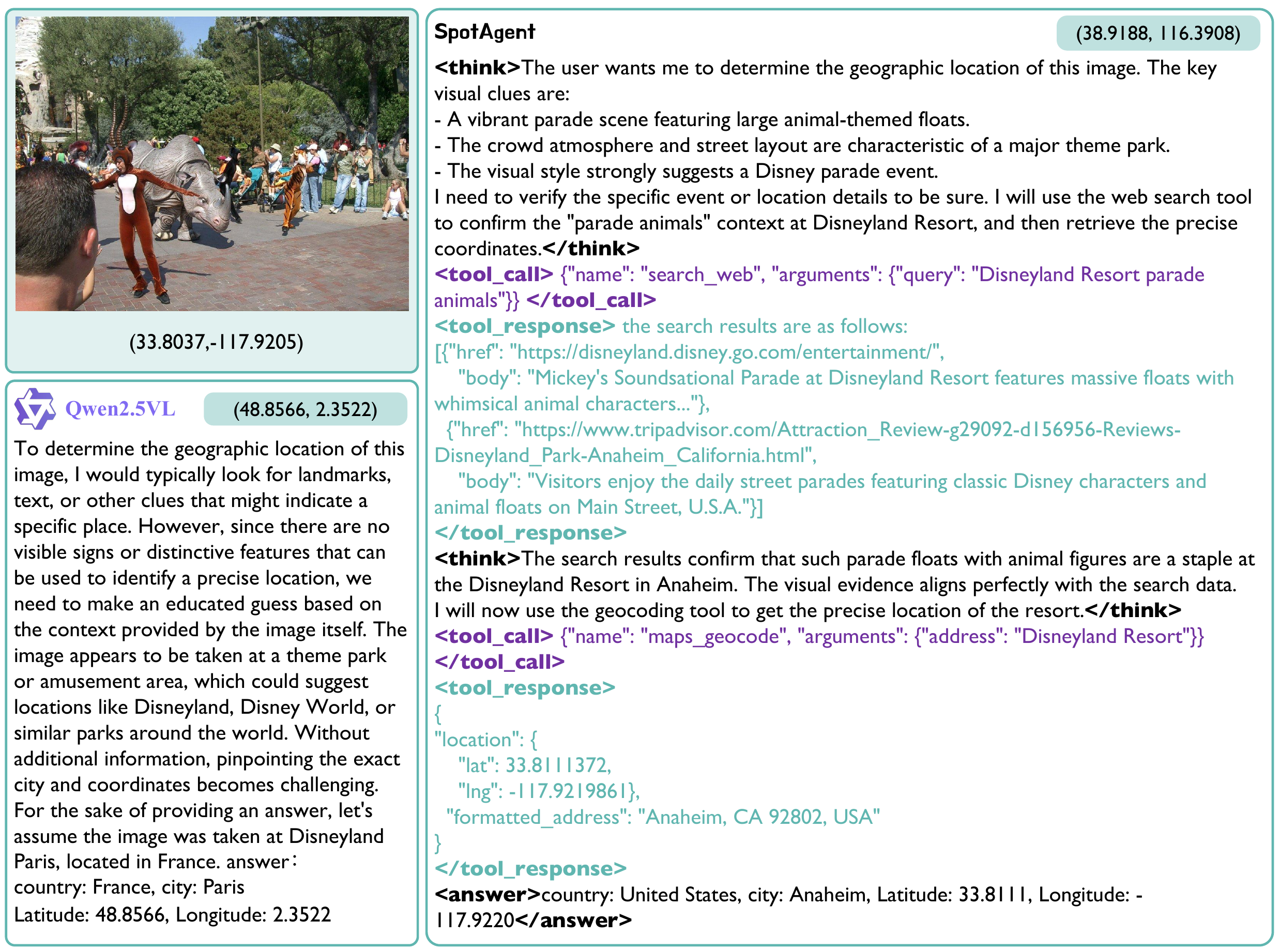}
    \caption{Qualitative comparisons with the base model.}
    \label{fig:app2}
\end{figure*}

\subsection{Im2GPS3k Benchmark}
The \textbf{Im2GPS3k} dataset~\cite{vo2017revisiting} serves as a standard evaluation protocol, originally curated to revisit the Im2GPS testbed in the deep learning era. Comprising approximately 3,000 query images collected from Flickr, it is distinct from training distributions that often exhibit bias towards popular tourist attractions. Instead, Im2GPS3k offers a balanced distribution of natural landscapes, urban settings, and iconic landmarks. This diversity necessitates that the model moves beyond simple pattern matching, requiring the ability to discern fine-grained visual cues (e.g., architectural styles, signage) and interpret broader geographic contexts (e.g., vegetation patterns, terrain) to achieve accurate localization on unseen data.

\subsection{YFCC4k Benchmark}
To assess the model's robustness and generalization on large-scale, in-the-wild data, we utilize the \textbf{YFCC4k} dataset, a curated subset derived from the Yahoo Flickr Creative Commons 100 Million (YFCC100M) collection~\cite{thomee2016yfcc100m}. This benchmark consists of 4,000 randomly sampled images that are strictly disjoint from the training set, representing a challenging ``hard'' evaluation setting. Characterized by the high entropy of user-generated content, YFCC4k includes a significant proportion of images with ambiguous visual features, varying illumination conditions, and diverse viewpoints. Consequently, high performance on this benchmark indicates a model's capacity for robust semantic reasoning rather than reliance on memorization of specific landmarks.
\subsection{Street View Text Dataset}
The \textbf{Street View Text} (SVT) dataset~\cite{wang2010word} was harvested from Google Street View to address the challenges of word recognition and spotting in unconstrained real-world environments. It comprises images of outdoor business signage and storefronts that exhibit variability in lighting conditions and viewpoints. 

\subsection{MP16-Pro Dataset}
In addition to standard benchmarks, we utilize the \textbf{MP16-Pro} dataset, an enhanced iteration of the original MP-16 introduced in  prior work~\cite{jia2024g3}. MP16-Pro augments these samples from standard MP-16 with hierarchical textual descriptions (e.g., ``Continent, Country, Region, City'') for every geo-tagged image. By filtering out samples with ambiguous or incomplete metadata while retaining the large-scale nature of the original dataset, MP16-Pro serves as a robust foundation for training multi-modal models.

\subsection{Evaluation Metrics}
Consistent with established protocols in prior literature~\cite{vo2017revisiting,vivanco2023geoclip}, we quantify geo-localization accuracy using the Great Circle Distance between the predicted location and the ground truth. Let $L_{pred} = (\phi_p, \lambda_p)$ and $L_{gt} = (\phi_g, \lambda_g)$ denote the latitude and longitude (in radians) of the predicted and ground truth locations, respectively. The geodesic distance $d(L_{pred}, L_{gt})$ represents the shortest path on the sphere and is calculated via the spherical law of cosines:
\begin{equation}
    d(L_{pred}, L_{gt}) = R \cdot \arccos\left( \sin\phi_p \sin\phi_g + \cos\phi_p \cos\phi_g \cos(\lambda_p - \lambda_g) \right)
\end{equation}
where $R \approx 6371$ km is the Earth's mean radius.

Based on this distance, we report the performance using the \textbf{Accuracy@D} metric, which measures the percentage of queries localized within a specific distance threshold $D$. Formally, the accuracy is defined as:
\begin{equation}
    \text{Accuracy}@D = \frac{1}{N} \sum_{i=1}^{N} \mathbb{I}(d(L_{pred}^{(i)}, L_{gt}^{(i)}) \le D)
\end{equation}
where $N$ denotes the total number of query images and $\mathbb{I}(\cdot)$ is the indicator function. To evaluate the model's reasoning capability across different scales, we report accuracy at a hierarchical set of thresholds: Street (1 km), City (25 km), Region (200 km), Country (750 km), and Continent (2500 km).

\subsection{Data Integrity and Contamination}
To ensure the integrity of our evaluation, we address a potential data leakage risk specific to the Im2GPS3k benchmark. The original images in this dataset are typically indexed or named using Flickr user IDs and associated metadata. With the integration of web search tools, a model could theoretically bypass visual reasoning by retrieving exact image pages or metadata directly from Flickr. To mitigate this ``shortcut'' and prevent data contamination, we implemented a strict anonymization pipeline: all original identifiers were systematically renamed into randomized hashes. This decoupling ensures that the search engine cannot pivot on original IDs, forcing the model to rely solely on visual cues and geographic knowledge rather than metadata retrieval.

\section{Detailed Results}

\subsection{More Qualitative Results}

\begin{figure*}[b]  
    \centering
    \includegraphics[width=0.84\linewidth]{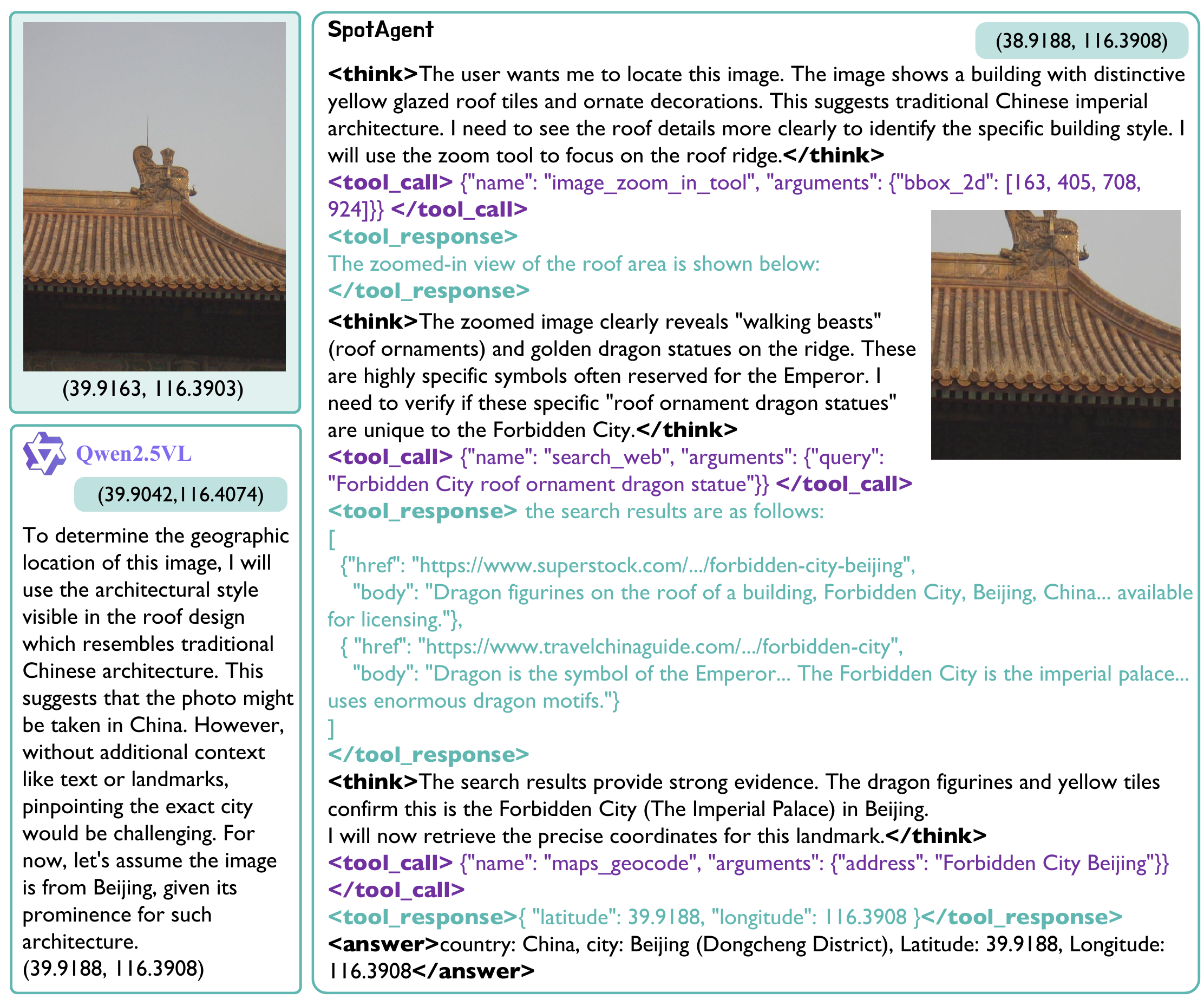}
    \caption{Qualitative comparisons with the base model.}
    \label{fig:app1}
\end{figure*}

More qualitative comparisons on the Im2GPS3k benchmark are shown in Figure~\ref{fig:app1} and Figure~\ref{fig:app2}.
In Figure~\ref{fig:app2}, while the standard CoT mode in the base LVLM correctly identifies the general context, it suffers from pinpointing exact locations due to a lack of external verification. In contrast, SpotAgent proactively invokes web search tools to cross-reference visual clues, such as specific parade floats, with real-world knowledge. By transforming obscure visual details into verifiable evidence, our agent successfully corrects the initial reasoning path and achieves far higher precision in complex, long-tail scenarios.

Figure~\ref{fig:app1} demonstrates SpotAgent's superiority in handling image with landmarks. 
While the base LVLM correctly identifies the general architectural style and predicts a reasonable location within Beijing, its lack of precise coordinate knowledge for the specific site results in a localization error of approximately 1.98 km.
In contrast, SpotAgent strategically employs the Image Zoom tool to inspect the walking beasts on the roof ridge. By cross-referencing these specific ornaments via Web Search, our agent transitions from a generalized regional guess to a pinpoint accurate localization of the Forbidden City with error of only 0.28 km.


\subsection{Performance of Generalist Open-Source LVLMs}

\begin{table}[htbp]
\centering
\small
\caption{Performance of generalist open-source LVLMs on the Im2GPS3k benchmark.}
\label{tab:vlm_comparison}
\setlength{\tabcolsep}{5pt} 
\begin{tabular}{@{}l ccccc@{}}
\toprule
 & \multicolumn{5}{c}{\textbf{Accuracy (\%)}} \\ 
\cmidrule(l){2-6} 
\textbf{Methods} & \textbf{1km} & \textbf{25km} & \textbf{200km} & \textbf{750km} & \textbf{2500km} \\
\midrule
InternVL3-8B   & 6.44 & 25.69 & 34.57 & 49.38 & 61.66 \\
Qwen2.5-VL-7B  & 3.97 & 25.48 & 43.65 & 64.29 & 77.90 \\
Qwen3-VL-8B    & 8.29 & 29.67 & 46.84 & 65.19 & 77.92 \\
InternVL3-78B  & 8.93 & 35.05 & 47.32 & 64.03 & 78.64 \\
Qwen2.5-VL-72B & 9.11 & 35.77 & 48.35 & 64.96 & 78.88 \\
Qwen3-VL-30B & \textbf{10.94} & \textbf{36.40} & \textbf{54.29} & \textbf{73.34} & \textbf{86.42} \\
\bottomrule
\end{tabular}
\end{table}

Results in Table \ref{tab:vlm_comparison} illustrates performances of some existing open-source LVLMs on the Im2GPS3k benchmark. 
For a fair comparison with prior specialized methods, SpotAgent is developed and trained based on the Qwen2.5-VL-7B backbone. This consistent setup ensures that the performance gains are primarily driven by our agentic reasoning and tool-use strategy rather than the base model's scale.

\clearpage
\subsection{Failure Cases of SpotAgent}

\begin{figure}[h]  
    \centering
    \includegraphics[width=\linewidth]{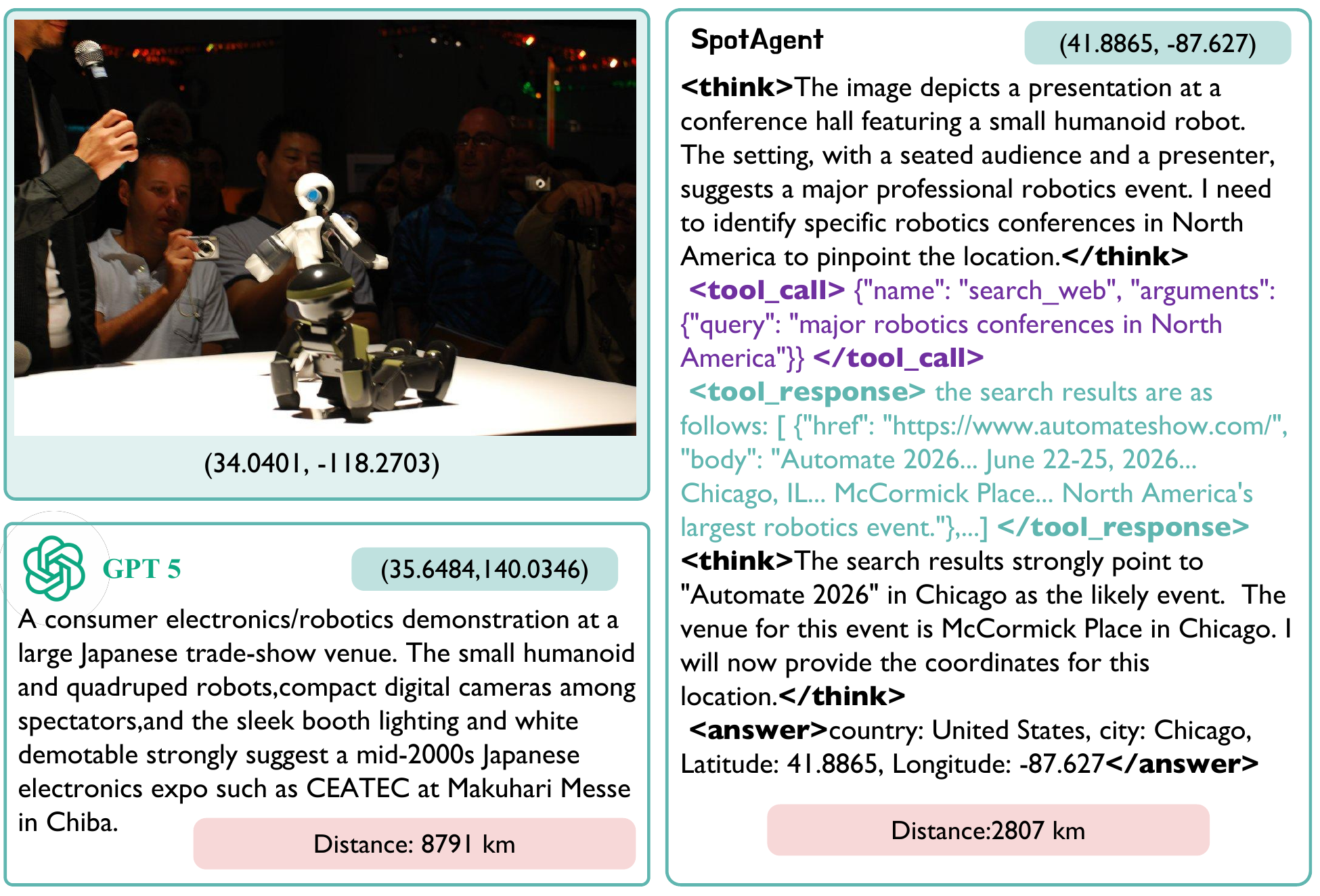}
    \caption{Failure case analysis on a spatially ambiguous indoor scenario.}
    \label{fig:bad1}
\end{figure}

There are challenges remain in handling generic indoor environments and spatially ambiguous events. As illustrated in Figure~\ref{fig:bad1}, scenarios such as conference halls, airports, or hotels often lack distinct, permanent architectural landmarks required for precise geo-grounding. In such cases, the agentic framework may correctly identify the type of activity (e.g., a robotics demonstration) but struggle to identify the specific location, as similar events occur globally.

\begin{figure}[h]  
    \centering
    \includegraphics[width=\linewidth]{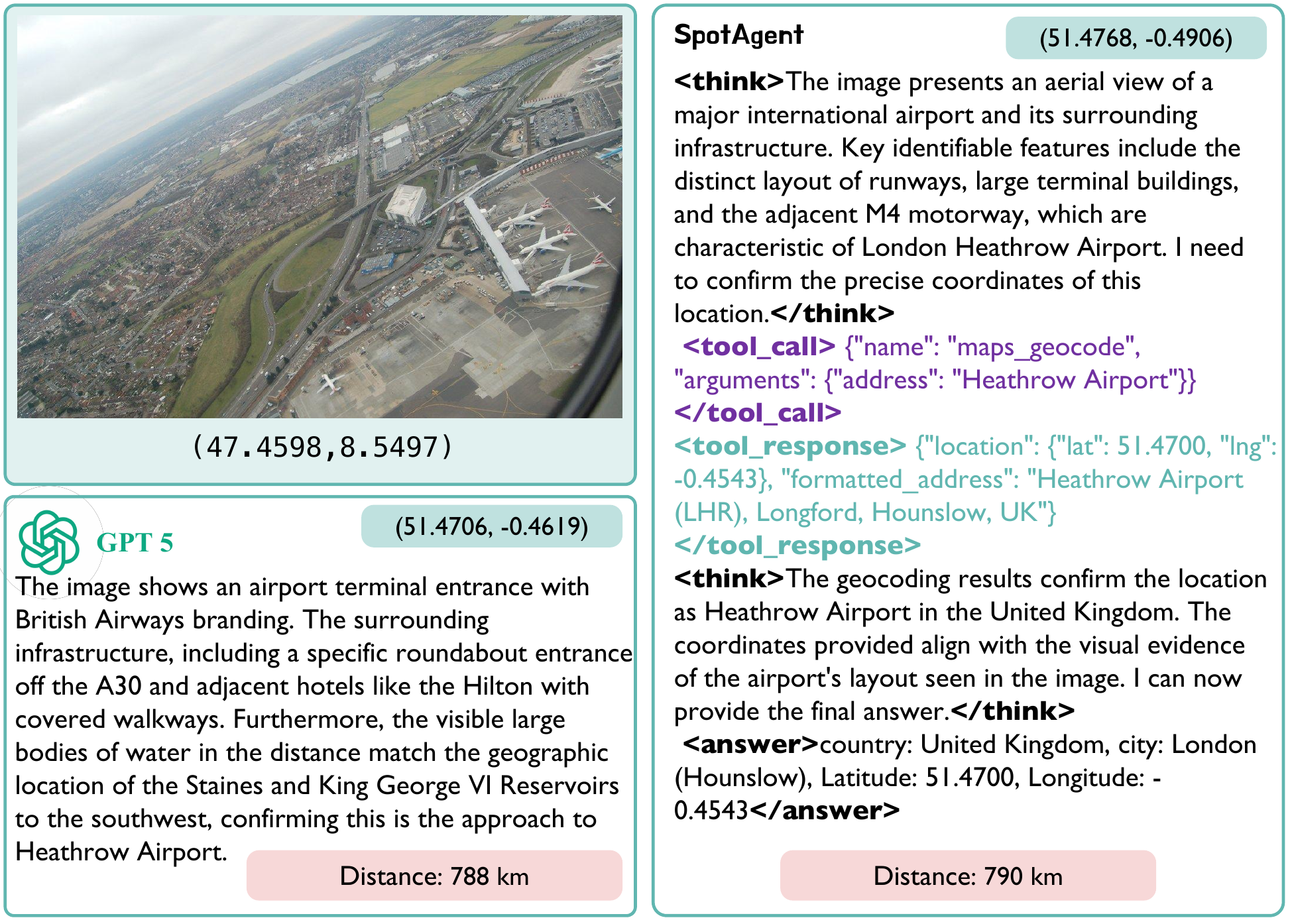}
    \caption{Failure case analysis on a confusing scene.}
    \label{fig:bad2}
\end{figure}

Figure~\ref{fig:bad2} presents a failure case where the model is deceived by strong, misleading semantic cues. The prominent British Airways livery on the tarmac acts as a visual trap, biasing the reasoning process toward the airline's primary hub (London Heathrow). Consequently, the agent ignores the specific geometric layout of Zürich Airport and instead hallucinates a spatial correspondence between the visible terrain, such as the configuration of nearby water bodies and roads, and the surroundings of Heathrow, highlighting the difficulty of decoupling instance-level localization from dominant semantic priors.

\section{Limitations and Future Work}
\label{limit}

The primary limitations of SpotAgent stem from the resource overhead required for data synthesis via commercial LLMs, as well as the operational costs incurred by real-time tool invocations during the inference process.
Future work can address this by developing a decentralized, agent-native information-seeking infrastructure to replace commercial APIs.
Another direction is to expand the toolkit, such as by integrating specialized image processing modules to further strengthen its fine-grained visual perception capabilities.
Furthermore, we can explore integrating agentic reinforcement learning to jointly optimize the tool-use policy. 

\end{document}